\documentclass[10pt,twocolumn,letterpaper]{article}

\usepackage{iccv}
\usepackage{times}
\usepackage{epsfig}
\usepackage{graphicx}
\usepackage{amsmath}
\usepackage{amssymb}
\usepackage{multirow}
\usepackage{booktabs}
\usepackage{caption}
\usepackage[accsupp]{axessibility}
\newcommand\blfootnote[1]{%
  \begingroup
  \renewcommand\thefootnote{}\footnote{#1}%
  \addtocounter{footnote}{-1}%
  \endgroup
}

\usepackage[pagebackref=true,breaklinks=true,letterpaper=true,colorlinks,bookmarks=false]{hyperref}

\iccvfinalcopy 

\def\iccvPaperID{7350} 

\ificcvfinal\fi

\begin{document}

\title{DnD: Dense Depth Estimation in Crowded Dynamic Indoor Scenes}

\author{Dongki Jung$^{\thanks{These two authors contributed equally.}}$\:\:\textsuperscript{\rm 1} \:\: Jaehoon Choi$^{\footnotemark[1]}$\:\:\textsuperscript{\rm 1,2} \:\: Yonghan Lee$^{1}$ \:\: Deokhwa Kim$^{1}$ \:\: Changick Kim$^{3}$ \\ 
Dinesh Manocha$^{2}$ \:\: Donghwan Lee$^{1}$ \\ 
$^{1}$NAVER LABS \:\: $^{2}$University of Maryland \:\: $^{3}$KAIST 
}

\vspace{-3mm}
\twocolumn[
{
\renewcommand\twocolumn[1][]{#1}%
\maketitle
\begin{center}
    \centering
    \includegraphics[width=0.85\linewidth]{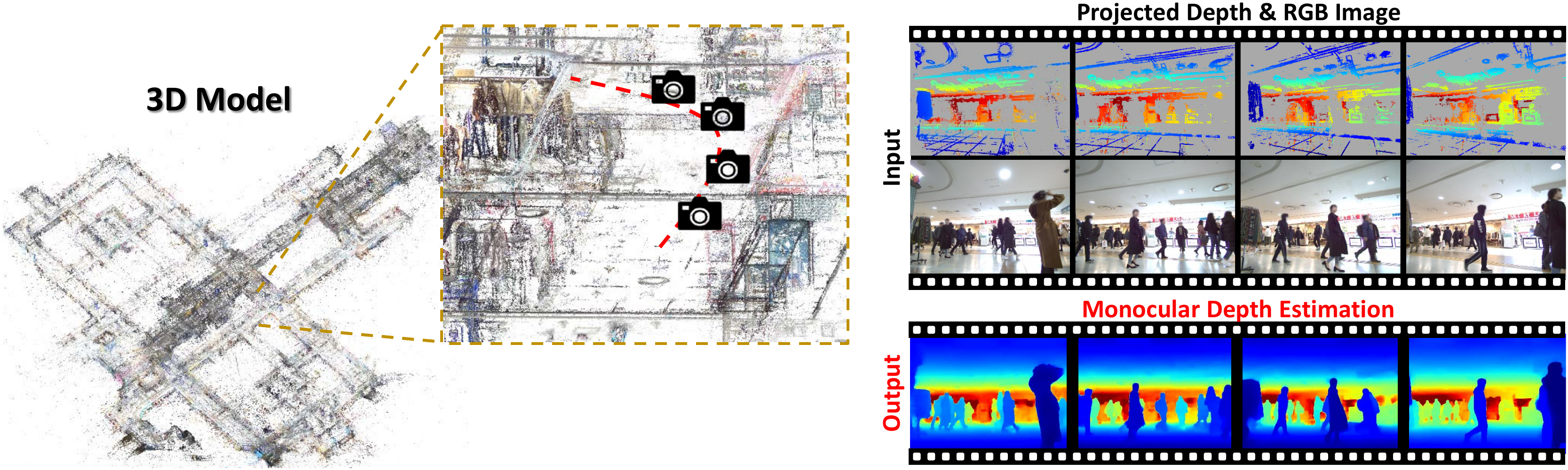}
    \vspace{-2mm}
    \captionof{figure}{With traditional 3D reconstruction methods \cite{schonberger2016structure,schonberger2016pixelwise}, we can obtain a 3D model on complex and crowded indoor environments. 
    Our novel approach, DnD, takes both an RGB image and a sparse depth map projected from this 3D model as input. 
    Our method predicts dense and absolute-scale depth maps of single view dynamic scenes.
    }
    \label{intro_figure}
    \vspace{-1mm}
\end{center}
}
]

\begin{abstract}
We present a novel approach for estimating depth from a monocular camera as it moves through complex and crowded indoor environments, e.g., a department store or a metro station. Our approach predicts absolute scale depth maps over the entire scene consisting of a static background and multiple moving people, by training on dynamic scenes. Since it is difficult to collect dense depth maps from crowded indoor environments, we design our training framework without requiring depths produced from depth sensing devices. Our network leverages RGB images and sparse depth maps generated from traditional 3D reconstruction methods to estimate dense depth maps.
We use two constraints to handle depth for non-rigidly moving people without tracking their motion explicitly. We demonstrate that our approach offers consistent improvements over recent depth estimation methods on the NAVERLABS dataset, which includes complex and crowded scenes. 
\end{abstract}

\section{Introduction}
\blfootnote{* These two authors contributed equally.}  
\blfootnote{Correspondence to dongki.jung@naverlabs.com, kevchoi@umd.edu}
There is considerable interest in using robotics and augmented reality technologies in crowded real-world spaces corresponding to malls, airports, or public places. In order to perform safe navigation or combine real and virtual worlds, robots \cite{liang2020vo,liang2020crowdsteer,sathyamoorthy2020densecavoid} or mobile devices \cite{stein2018learning,lynen2020large} need a 3D geometric representation of large-scale indoor environments. While there is considerable progress in terms of capturing depth using LiDARs or stereo cameras, existing devices still have their own limitations. For example, 3D LiDARs \cite{christian2013survey} tend to produce sparse depth maps for distant objects and may result in noisy point cloud maps due to the high level of occlusions caused by multiple moving people. Moreover, because of the high prices and large volumes of the depth sensors, there is a critical need to consider the case where only a single camera is available.  

Given a large number of video frames, traditional 3D reconstruction methods such as structure-from-motion (SfM) and multi-view stereo (MVS)  \cite{furukawa2015multi,snavely2008modeling,schonberger2016structure,schonberger2016pixelwise} can generate a 3D model. Recently, visual localization techniques \cite{piasco2018survey} have been developed to allow mobile devices to obtain the location and camera pose from which the image is taken. Given the 3D model and visual position, however, mobile devices moving through crowded indoor environments are only able to capture sparse and highly noisy depth maps. 
This is because traditional reconstruction methods are based on both a static scene assumption (that static areas of the scenes can be observed from two different viewpoints) and correct feature matching in two or more images. However, the moving pedestrians in crowded indoor environments tend to violate the static scenes assumption. Moreover, traditional 3D reconstruction methods may fail to perform the correct matching on non-textured (e.g., walls), specular, and reflective regions (e.g., glass) of the scene. Consequently, these problems lead to a 3D model on complex and crowded indoor environments.           

\noindent\textbf{Main Results:} 
To address these limitations, we explore new methods that can utilize the 3D model generated using traditional 3D reconstruction methods \cite{schonberger2016structure,schonberger2016pixelwise} into learning-based depth estimation algorithms for dynamic scenes. Given the 3D model, our method can be used for general applications because it can compute dense depth maps of dynamic scenes without reconstructing this 3D model iteratively.   
In contrast to supervised learning methods \cite{li2018megadepth,chen2019learning,li2019learning}, our method does not rely on dense depth maps generated from depth sensing devices.
Our approach, shown in Fig.\:\ref{intro_figure}, takes an RGB image and a sparse depth map projected from the 3D model as input and outputs a dense depth map. Given the pose obtained from the SfM \cite{schonberger2016structure}, we propose using the photometric consistency loss, which enables our method to estimate dense depth maps, and depth loss, forcing our network to learn absolute-scale depth. Although these loss functions are useful for providing dense depth maps in static background regions, there still exist great challenges in estimating depths of multiple non-rigidly moving objects, i.e. the pedestrians. To overcome this limitation, we propose two constraints: 1) a flow-guided shape constraint to refine the depth maps for human regions by filling missing parts of the human regions and removing visual artifacts, and 2) a normal-guided scale constraint to force our neural network to learn the absolute scale depth in human regions guided by depth values in the human's ground contact point. 

Compared to traditional reconstruction approaches or recent learning-based methods \cite{li2019learning,yu2020fast,lee2019big}, our approach (DnD) shows a better ability to predict plausible depth in both human and non-human regions, though we only use a monocular camera. We evaluate our approach on the NAVERLABS dataset \cite{lee2021large}, the first dataset that provides both metric 3D SfM models and dynamic scenes collected from a department store and a metro station.
The main contributions of this paper are summarized as follows: 
\begin{itemize}
    \item We introduce a novel approach to estimate the depth maps using both dynamic scenes collected from a moving monocular camera and given sparse depth maps. 
    We train the monocular depth estimation network with a 3D model generated by
    traditional 3D reconstruction algorithms \cite{schonberger2016structure,schonberger2016pixelwise}.   
    \item We present two novel constraints based on optical flow and surface normal that improve the accuracy of our monocular depth estimation network to predict absolute scale depths for moving people.    
    \item 
    We highlight the benefits of our approach over the state-of-the-art methods on crowded indoor environments and observe 3.6\%\:-10.2\% improvement in RMSE. Furthermore, our method works well in diverse indoor datasets like TUM RGB-D and NYUv2.
\end{itemize}


\section{Related Work}

\noindent\textbf{Structure from Motion and Multi-view Stereo}
Traditional SfM systems \cite{snavely2008modeling,agarwal2011building,schonberger2016structure} rely on matching features across two or more images of the same scene and using epipolar geometry \cite{hartley2003multiple} to reconstruct depth. Multi-view stereo (MVS) algorithms estimate the dense 3D structure of a scene with multiple calibrated images from the arbitrary viewpoints \cite{furukawa2015multi}. Recently, some researchers have presented learning-based MVS methods \cite{yao2018mvsnet,yu2020fast} that build on a neural network to learn the regularization of 3D cost volume. 
However, traditional SfM and MVS methods often produce sparse and erroneous 3D reconstruction due to incorrect feature matches and dynamic objects \cite{ozyesil2017survey}.
Due to static scene assumption, they either drop pixels with low confidence or estimate incorrect depth values for dynamic objects.   
 

\noindent\textbf{Monocular Depth Estimation}
The existing learning-based depth estimation methods fall into two groups. One group is based on supervised learning \cite{eigen2014depth,eigen2015predicting,liu2015learning,laina2016deeper,fu2018deep,lee2019big,yin2019enforcing}, which requires a large-scale dataset with groundtruth depth maps. However, all of these methods require dense groundtruth collected from active depth sensors. 
Other recent works explore the idea of training with weak supervisions, e.g., using ordinal depth relations as groundtruth \cite{zoran2015learning,chen2016single,xian2018monocular,xian2020structure} and leveraging multi-view stereo reconstruction algorithms to generate pseudo groundtruth from internet photo collections \cite{li2018megadepth,chen2019learning}. The other group is based on self-supervised learning using either rectified stereo image pairs \cite{garg2016unsupervised,godard2017unsupervised} or monocular video sequences \cite{zhou2017unsupervised,yin2018geonet,godard2019digging,guizilini20203d,xue2020toward, yang1711unsupervised,choi2020safenet} as training data. 
Video-based depth estimation methods \cite{liu2019neural,Teed2020DeepV2D,luo2020consistent} use temporally consecutive frames to estimate depth over time during inference under the assumption of handling only static scenes.
Several works for monocular depth completion \cite{eldesokey2019confidence,ma2019self,xu2019depth,yang2019dense,SelfDeco} have been proposed to capitalize on sparse depth maps with corresponding images, resulting in dense depth estimations. 

\begin{figure*}[t]
    \centering
    \includegraphics[width=0.8\linewidth]{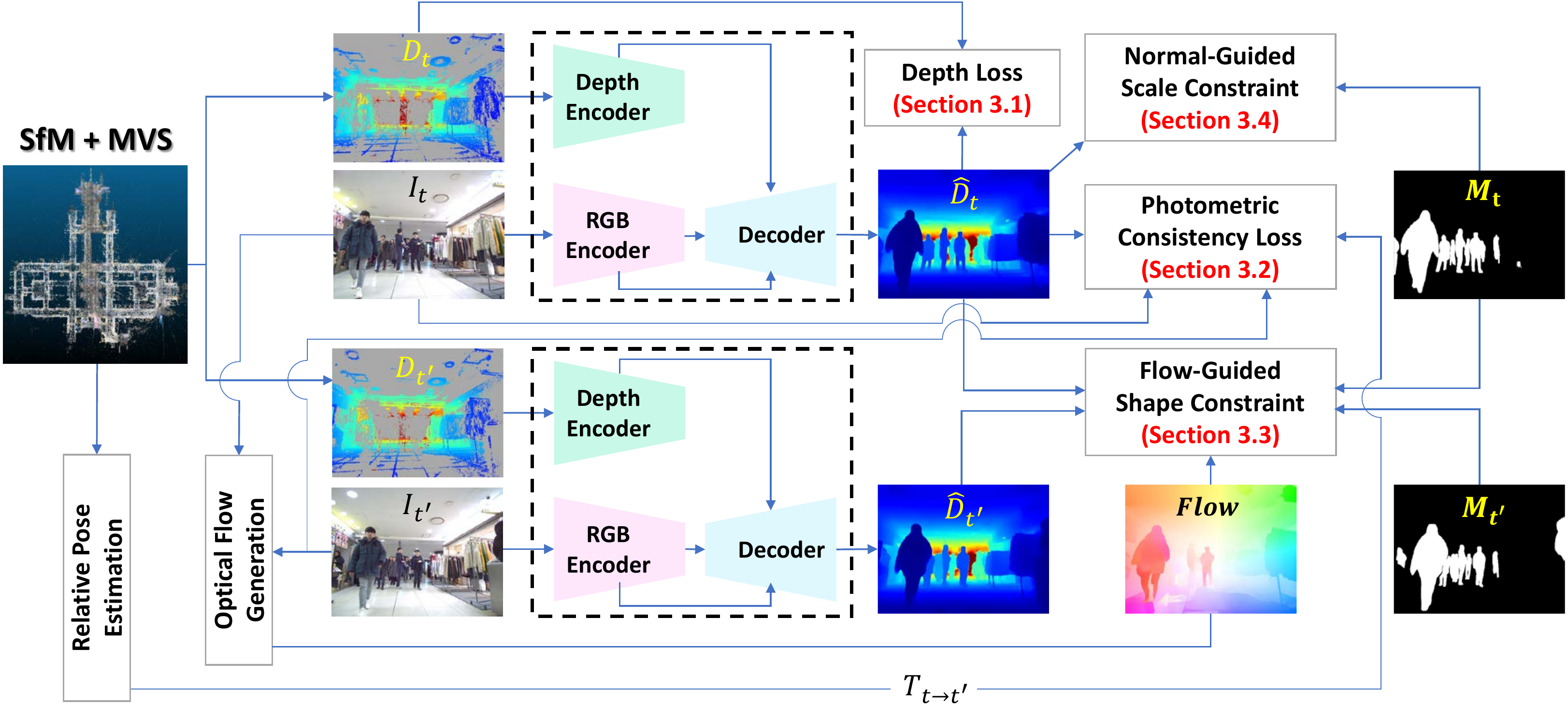}
    \vspace{-2mm}
    \caption{Overview of our proposed approach. The black box with a dotted line shows our monocular depth estimation network consisting of an RGB encoder, a depth encoder, and a decoder. For training, the network takes consecutive temporal frames ($I_{t}$, $I_{t^{\prime}}$) with corresponding sparse depth maps ($D_{t}$, $D_{t^{\prime}}$) projected from the 3D model as input. 
    Our training method is a combination of four terms. 1) The depth loss encourages our network to encode absolute scale from a depth input. 2) The photometric consistency loss is based on view synthesis and regularizes the network training for static background regions. 3) The flow-guided shape constraint enables the network to complete missing pixels in human regions with proper depth values and eliminate visual artifacts. 
    4) The normal-guided scale constraint enables our network to estimate accurate and absolute-scale depths on moving people. 
}
    \label{fig:framework}
    \vspace{-4mm}

\end{figure*}

\noindent\textbf{Depth Estimation of a Dynamic Scene}
Existing works \cite{russell2014video,ranftl2016dense} use object-level motion segmentation to reconstruct a dynamic scene. 
The key challenge of applying deep neural networks to this task is the lack of large-scale datasets containing diverse and dynamic scenes. Many works adopt a data-driven approach by building diverse datasets from either internet stereo videos \cite{wang2019web} or 3D videos \cite{ranftl2020towards}. Some works \cite{li2019learning,chen2019learning} use SfM and MVS to build a dataset with depth groundtruth from internet video collections. Li et al. \cite{li2019learning} use a collection of Mannequin Challenge videos to train a network for depth estimation in dynamic scenes. Yoon et al. \cite{yoon2020novel} present a view synthesis of a dynamic scene via monocular depth estimation with a pre-trained model \cite{ranftl2020towards} and MVS. 
Luo et al. \cite{luo2020consistent} finetune a pre-trained model to satisfy 3D geometric constraints on consecutive video frames. Although their method can handle scenes with a moderate object motion, it is vulnerable to crowded scenes with extreme object motion. In comparison, our method does not require either particular datasets or the pre-trained depth estimation network. Our approach can be generalized to large-scale datasets in real-world crowded indoor environments.  

\section{Our Method: DnD}

We present a learning-based approach to estimate dense and absolute-scale depth maps in scenes from complex and crowded indoor environments. Intuitively, most scenes consist of a static background (e.g., a wall) and dynamic objects (e.g., moving people). The sparse depth maps projected from the 3D model have absolute-scale depth values in small regions of the static background. The rest of the regions in the projected depth maps, including the static background and dynamic objects, have empty depth values. 
Our proposed approach trains a dense depth estimation network from monocular video sequences with the corresponding sparse depth maps. 
As shown in Fig.\:\ref{fig:framework}, our depth estimation model takes a current image $I_{t}$ with a sparse depth map $D_{t}$ and temporally adjacent images $I_{t^{\prime}}$ with sparse depth maps $D_{t^{\prime}}$. The sparse depth maps $D_{t}$ and $D_{t^{\prime}}$ are projected from the 3D model. The $I_{t^{\prime}}$ includes two temporal frames \(I_{t-1}\) and \(I_{t+1}\). Our depth estimation model predicts dense depth maps $\hat{D}_{t}$ and $\hat{D}_{t^{\prime}}$ to compute a flow-guided shape constraint.  

\subsection{Absolute-Scale Depth Loss}
\label{subsec:absolute-scale}
Generally, monocular depth estimation methods suffer from an inherent scale ambiguity problem. To mitigate this limitation, we use an absolute-scale depth input $D_{t}$ as a groundtruth.    
We apply the L1 loss to penalize the differences between the depth input $D_{t}$ and the depth prediction $\hat{D}_{t}$ on the pixels where $D_{t}$ values exist. The depth loss is formulated as,       
\begin{equation} \label{}
    L_{d} = \sum_{p \in \Omega }{\parallel \hat{D}_{t}(p) - D_{t}(p) \parallel},
\end{equation}
where $\Omega$ indicates valid points that involve available sparse depths. This loss enables our network to learn absolute-scale depth values in the regions of the depth input and further extrapolate depth values with absolute scales in empty regions including a static background and moving people.

\subsection{Photometric Consistency Loss}
\label{subsec:photometric}
To train our network, we turn to previous self-supervised monocular depth estimation methods \cite{zhou2017unsupervised,godard2019digging} that have utilized the photometric loss as the main loss function for training the network. 
We are able to generate a synthesized frame $I_{t}^{\prime}$ by reprojecting temporally adjacent frames $I_{t^{\prime}}$ to the current frame $I_{t}$ with the camera intrinsic matrix $K$, the predicted depth $\hat{D}_{t}$, and the relative pose $T_{t \rightarrow t^{\prime}}$. This relative pose between consecutive temporal frames is computed from the absolute poses from SfM \cite{schonberger2016structure}.
The photometric consistency loss with a combination of L1 and SSIM \cite{wang2004image} is formulated as follows:
\begin{equation} \label{photometric}
\begin{split}
    L_{ph} = \alpha& \frac{1 - \text{SSIM}(I_{t}, I^{\prime}_{t})}{2} 
    + (1 - \alpha) \parallel I_{t} - I_{t}^\prime\parallel,\\
    I_{t}^{\prime}(p)  &= I_{t^{\prime}}\langle\pi (K T_{t \rightarrow t^{\prime}} \hat{D}_{t}(p) K^{-1} \tilde{p})\rangle,
\end{split}
\end{equation}
where $\alpha = 0.85$, $\tilde p$ is the homogeneous coordinate of $p$, $\pi$ means projection from homogeneous to image coordinates, and $\langle \cdot \rangle$ indicates the bilinear sampling function. Note that one current frame and two temporally adjacent frames are used to compute the photometric consistency loss. Following \cite{godard2019digging}, we remove the occluded and out-of-view pixels, and also mask out low texture regions via minimum reprojection loss and auto-masking techniques. In particular, this photometric consistency loss encourages our network to predict dense depth maps over the static backgrounds.     

\subsection{Flow-Guided Shape Constraint} 
\label{subsec:Flow-Guided}

The loss functions described in the previous sections are still limited in their ability to provide accurate depths when there are moving people. This is because both MVS and photometric consistency loss operate under the assumption of a single moving camera and a static scene. This assumption implies that inconsistencies between different views of each image are only derived from the camera ego-motion in static scenes  \cite{hartley2003multiple}. If we correctly estimate the motion of moving people between the consecutive frames, the view consistency for 3D reconstruction can be obtained for two different views. However, with multiple moving people in dynamic scenes, triangulation-based methods fail to achieve a scene consistent depth map. Since multiple moving people in scenes undergo non-rigid deformations, it is difficult to explicitly estimate their 3D motion \cite{wang2019web,li2019learning,kumar2019dense}. 

Instead of modeling object motions in 3D, we leverage monocular video frames to estimate temporally coherent depth maps with respect to moving camera motions and non-rigidly moving people. To achieve this, we use optical flow $F$, a human mask $M_{t}$ from the current frame $I_{t}$, and human masks $M_{t^{\prime}}$ from the temporally adjacent frames $I_{t^{\prime}}$. 
Given consecutive frame pairs, the optical flow describes which pixel pairs have the same intensities. Human masks are also used to find correct regions that overlap due to the high-level of occlusion, non-rigid deformation, and complex ego-motions of the moving camera.             
\begin{equation} \label{eq:humanmask}
    M = M_{t} \cap F_{t^{\prime} \rightarrow t}(M_{t^{\prime}}),
\end{equation}
where $M$ is the overlapped region of the flow-warped human mask $M_{t^{\prime}}$ and the current human mask $M_{t}$ to prevent occlusion issues, meaning that some pixels from $M_{t}$ do not appear in $F_{t^{\prime} \rightarrow t}(M_{t^{\prime}})$. 
In dynamic scenes captured by the moving camera, the absolute depth values for dynamic human regions can be inconsistent. Thus, we apply the optical flow field to compare consecutive inverse depth predictions for human regions in gradient domains. Our proposed flow-guided shape constraint is defined as
%
\begin{equation} \label{flow-guided}
    \begin{split}
    \overline{\triangledown}(d^{*}&(p)) = \frac{\triangledown{d^{*}(p)}}{\mid d^{*}(p)+\triangledown{d^{*}(p)}\mid + \mid d^{*}(p) \mid},\\
    L_{f} &= \frac{1}{\mid M \mid}\sum_{p \in M} \mid \overline{\triangledown}{d^{*}_{t}(p)} - \overline{\triangledown}{F_{t^{\prime} \rightarrow t}(d^{*}_{t^{\prime}}(p))} \mid
    \end{split}
\end{equation}
where $\triangledown$ and $\overline{\triangledown}$ denote the gradient and scale-invariant gradient and $d^{*}_{t} = d_{t} / \mu(d_{t})$ indicates the mean-normalized inverse depth. Our flow-guided shape constraint enforces smooth gradients and completes missing parts of human regions with accurate depth values. To regularize the depth in a static background, we impose the edge-aware depth smoothness loss $L_{s}$ over pixels in non-human regions $M^{C}$:
\begin{equation} \label{}
    L_{s} = \sum_{p \in M^{C}}\mid\triangledown{d^{*}_{t}}(p)\mid e^{-\mid\triangledown{I_{t}(p)}\mid}.
\end{equation}

\subsection{Normal-Guided Scale Constraint}
\label{subsec:Normal-Guided}
\begin{figure}[t]
    \centering
    \includegraphics[width=\linewidth]{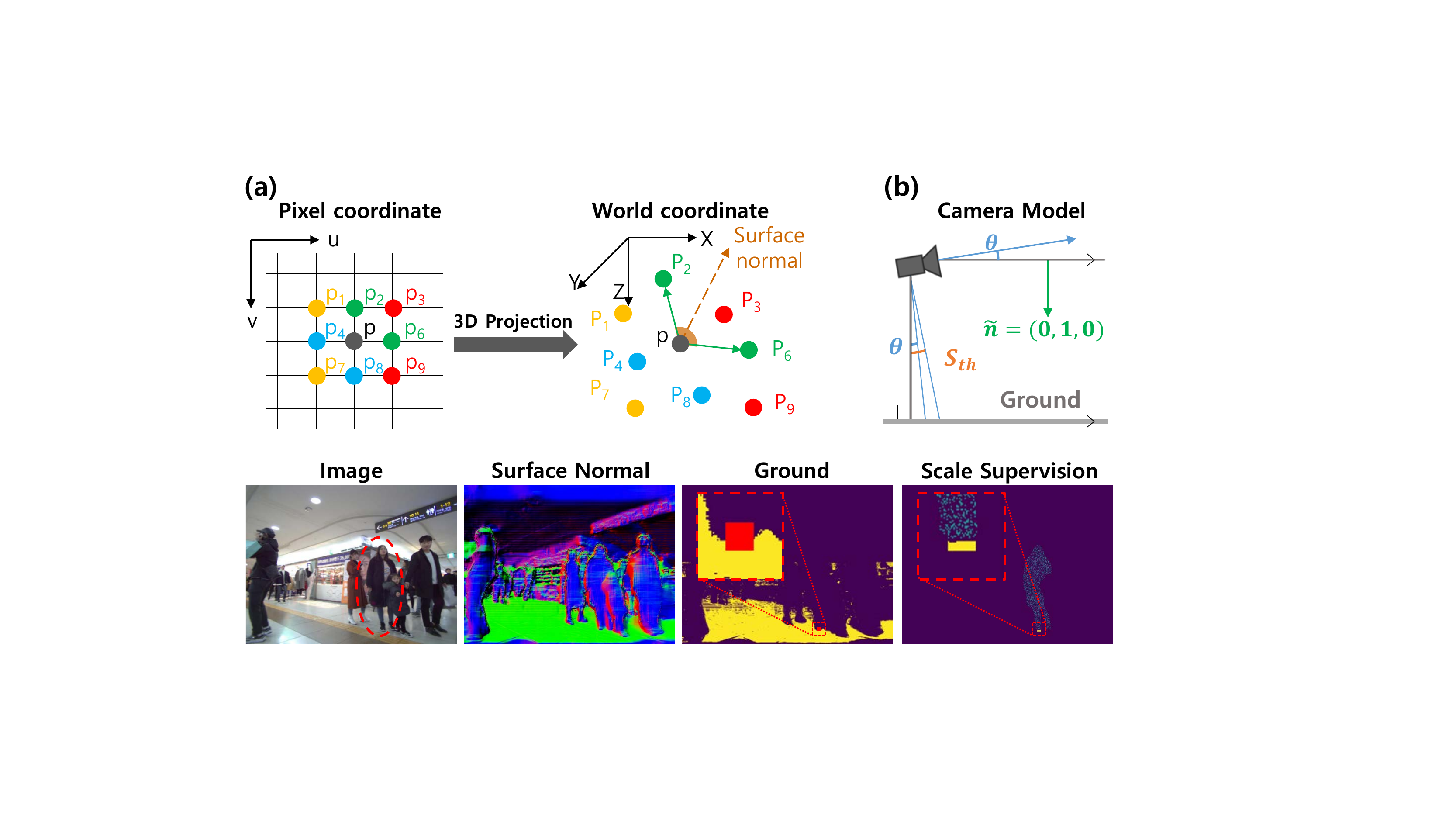}
    \vspace{-5mm}
    \caption{(a) The process to obtain a surface normal from the pixel coordinates and the corresponding depth values. (b) shows that the ground normal $\tilde{n}$ is not aligned with the y-direction of the camera coordinates.
    The bottom images indicate the process for propagating the reliable scale for the person in the dashed red circle.
    The red point in the third column means the small patch $B$, indicating the human's ground contact point. In the fourth column, the yellow and blue points show the ground-patch intersection and uniformly sampled points in the human mask respectively.}
    \label{fig:surface normal}
    \vspace{-3mm}
\end{figure}
To achieve scale-aware depth estimation, accurate depth estimation with an absolute scale for moving people is crucial.
We leverage absolute-scale depth values in the static background to constrain the scale of depth values for moving people. 
Intuitively, people are standing on the ground and the depth values of the humans are nearly the same as the ones on the ground. The human's ground contact point can be a significant geometric cue for scale-aware depth estimation. 
Thus, we introduce weak supervision, which forces several randomly selected points corresponding to the particular human mask to be consistent with the ground contact points' depth values.

Inspired by \cite{yang1711unsupervised,xue2020toward}, we estimate the ground area using the surface normal.
The 8-neighbors convention is used to specify the normal direction, as seen in Fig.\:\ref{fig:surface normal} (a). Those 8 points at the 2D coordinate are split into 4 pairs, where each pair of vectors is perpendicular, e.g., $(i,j) = \{(2,6), (3,9), \cdots \}$. The average of the four normals determines the final surface normal for the position $p$:
\begin{equation} \label{}
\begin{split}
    N_{t}(p) = \frac{1}{4}\sum_{\overrightarrow{pp_{i}} \perp \overrightarrow{pp_{j}}}\frac{\overrightarrow{PP_{i}} \times \overrightarrow{PP_{j}}}{\parallel \overrightarrow{PP_{i}} \times \overrightarrow{PP_{j}} \parallel}_{2}, \\
\end{split}
\end{equation}
where $P$ means the reconstructed 3D point clouds for pixel $p$ by the predicted depth and the operator $\times$ denotes the cross product.
Among many pixel points $p$, we only need points in ground regions.   
The true normal direction of ground $\tilde{n}=(0,1,0)$ and the estimated normal $N_{t}(p)$ should be matched.
As in Fig.\:\ref{fig:surface normal} (b), the camera is not perpendicular to the ground by a difference of $\theta$ for the camera model, and there exists the uncertainty of the estimated normal $N_{t}(p)$.
Thus, we set $S_{th}$ as a threshold to find the ground area and calculate the cosine similarity between the true ground normal $\tilde{n}$ and the predicted normal $N_{t}(p)$:
\begin{equation} \label{}
    G = \{p \mid \:\: \mid cos^{-1}\frac{\tilde{n} \cdot N_{t}(p)}{\parallel \tilde{n} \parallel \parallel N_{t}(p) \parallel} \mid \:<\: S_{th} \},
\end{equation}
where the operator $\cdot$ denotes the inner product.
Then, in order to determine the human's ground contact point, we detect small patches $B$ centered at the bottom of the pixel in the human instances. These pixels in both the small patches and the ground area are possible candidates for the human's ground contact point.
However, simply applying the constant value as supervision risks severe degradation of depth values in the human region because moving people have exquisite and isometric shapes.
Therefore, we uniformly sample the pixels $M^{\prime}$ in the human mask $M_{t}$ and constrain them by the median value in the intersection of $B$ and $G$:
\begin{equation} \label{eq:weqk_supervision}
    L_{n} = \sum_{p \in M^{\prime}}\frac{{\mid \hat D_{t}(p) - med(\hat{D}_{t}(B \cap G)) \mid}}{\hat D(p)},
\end{equation}
where the operator $med$ takes the median value of factors.

\subsection{Overall Loss Function}
Our overall loss function is a weighted sum over the previously introduced losses,
\begin{equation}
    L_{total} = \lambda_{d}L_{d} + \lambda_{p}L_{ph} + \lambda_{s}L_{s} + \lambda_{f}L_{f} + \lambda_{n}L_{n}
\label{total_loss}
\end{equation}
where \(\lambda_{d},\lambda_{p},\lambda_{s},\lambda_{f}\) and \(\lambda_{n}\) denote weights on the respective loss terms selected through a grid search.

\section{Experiments}
\subsection{Experiment Settings}
To validate our method, we experiment with the NAVERLABS Dataset, NYUv2 and TUM RGB-D dataset. The depth value is represented in the metric scale (m). We evaluate our method using the standard metrics \cite{eigen2014depth}.

\begin{table*}[t]
    \centering
    \resizebox{0.9\textwidth}{!}
    {
    \begin{tabular}{l|c|c|cccc|ccc}
    \toprule
    \multirow{2}{*}{Method} & \multirow{2}{*}{Data} & \multirow{2}{*}{Scaling} & \multicolumn{4}{c|}{F+B / F  ( Lower is better )} & \multicolumn{3}{c}{F+B / F  ( Higher is better )} \\
    & & &  Abs Rel & Sq Rel & RMSE & RMSE log & $\delta_{1.25}$ & $\delta_{1.25^2}$ & $\delta_{1.25^3}$ \\     
    \hline
    Fast-MVSNet \cite{yu2020fast} & MS & - & 0.383 / 0.754 & 0.60 / 1.37 & 2.05/ 2.95 & 0.123/ 0.213 & 0.634 / 0.358 & 0.787 / 0.547 & 0.866 / 0.697 \\
    DepthComple \cite{ma2019self} & MS & - & 0.800 / 0.863 & 0.7 / 0.79 & 4.52 / 4.67 & - / - & 0.041 / 0.024 & 0.087 / 0.050 & 0.139 / 0.083 \\
    MiDaS \cite{ranftl2020towards} & MS & median & 0.413 / 0.459 & 0.33 / 0.42 & 2.77 / 2.86 & 0.181 / 0.191 & 0.380 / 0.352 & 0.627 / 0.604 & 0.785 / 0.773 \\
    MC \cite{li2019learning} & MS & median & 0.294 / 0.315 & 0.30 / 0.22 & 2.05 / 2.53 & 0.122 / 0.162 & 0.585 / 0.459 & 0.822 / 0.741 & 0.906 / 0.847 \\
    BTS \cite{lee2019big} & MS & - & 0.396 / 0.696 & 0.66 / 1.24 & 1.96 / 2.70 & 0.120 / 0.198 & 0.665 / 0.401 & 0.792 / 0.589 & 0.862 / 0.726 \\
    \midrule
    DnD ($L_{ph}$ \text{only}) & MS & median & 0.326 / 0.596 & 1.38 / 4.16 & 2.39 / 3.60 & 0.107 / 0.177 & 0.711 / 0.545 & 0.840 / 0.705 & 0.898 / 0.795 \\
    DnD & MS & - & \textbf{0.189} / \textbf{0.240} & \textbf{0.20} / \textbf{0.16} & \textbf{1.76} / \textbf{2.44} & \textbf{0.084} / \textbf{0.133} & \textbf{0.806} / \textbf{0.677} & \textbf{0.881} / \textbf{0.798} & \textbf{0.919} / \textbf{0.856} \\
    \midrule
    Fast-MVSNet \cite{yu2020fast} & Dept & - & 0.551 / 0.889 & 0.98 / 1.97 & 3.24 / 4.11 & - / 0.243 & 0.431 / 0.288 & 0.655 / 0.498 & 0.791 / 0.654 \\
    DepthComple \cite{ma2019self} & Dept & - & 0.842 / 0.897 & 0.76 / 0.84 & 5.88 / 6.65  & - / - & 0.014 / 0.058 & 0.028 / 0.028 & 0.093 / 0.046 \\
    MiDaS \cite{ranftl2020towards} & Dept & median & 0.461 / 0.519 & 0.46 / 0.54 & 4.39 / 3.72 & 0.208 / 0.210 & 0.375 / 0.332 & 0.579 / 0.554 & 0.729 / 0.728 \\
    MC \cite{li2019learning} & Dept & median & 0.428 / 0.385 & 0.40 / \textbf{0.29} & 3.91 / 3.36 & 0.190 / 0.178 & 0.364 / 0.384 & 0.626 / 0.676 & 0.786 / 0.822 \\
    BTS \cite{lee2019big} & Dept & - & 0.584 / 1.066 & 1.36 / 2.80 & 3.06 / 4.40 & 0.159 / 0.260 & 0.561 / 0.327 & 0.721 / 0.488 & 0.808 / 0.614 \\
    \midrule
    DnD ($L_{ph}$ \text{only}) & Dept & median & 0.289 / 0.388 & 0.47 / 0.60 & 2.60 / 3.24 & 0.109 / 0.148 & 0.663 / 0.564 & 0.837 / 0.759 & 0.911 / 0.847 \\
    DnD  & Dept & - & \textbf{0.213} / \textbf{0.250} & \textbf{0.32} / 0.30 & \textbf{2.36} / \textbf{3.04} & \textbf{0.084} / \textbf{0.116} & \textbf{0.761} / \textbf{0.707} & \textbf{0.889} / \textbf{0.836} & \textbf{0.932} / \textbf{0.886} \\
    \bottomrule
    \end{tabular}
    }
    \vspace{-2mm}
    \caption{Quantitative comparisons with the state-of-the-art depth estimation algorithms. F means the evaluation results in the human regions and F+B indicates the evaluation results on depth values over the entire scene. MS and Dept denote the Metro Station dataset and the Department Store dataset, respectively. 
    DnD ($L_{ph}$ only) is trained only using the photometric consistency loss. In the MS dataset, DnD shows improvement in depth for human regions by 3.6\% in the RMSE metric. In terms of depth for entire regions, DnD reduces the RMSE by 10.2\%.  
    }
    \label{table: metric}
    \vspace{-3mm}
\end{table*}

\noindent\textbf{NAVERLABS Indoor Localization:} The NAVERLABS dataset \footnote{is available at https://www.naverlabs.com/datasets} \cite{lee2021large} comprises scenes collected from two different places: Department Store (Dept) and Metro Station (MS). 
This dataset is split into 60K images for training and 835 images for testing in MS, and 25K images for training and 443 images for testing in Dept.
To evaluate the crowdedness of indoor datasets, we use a crowd density, which is a ratio between the area of human pixels in scenes and the area of all image pixels. Most public indoor datasets \cite{silberman2012indoor,armeni2017joint,dai2017scannet,chang2017matterport3d,shotton2013scene,taira2018inloc} 
do not have scenes with moving people (\textless\: 0.1\% crowd density). 
In NAVERLABS, Dept and MS have 6.87\% and 12.9\% crowd density, the most crowded dataset. 
Additionally, Dept and MS have 6.7 and 3.6 people per scene on average. Different types of datasets \cite{chen2016single,xian2018monocular,wang2019web,li2019learning} that include diverse dynamic scenes contain only images or ordinal depth relations for pairs of points. 

NAVERLABS used all the images collected from 6 cameras to build a 3D model via COLMAP \cite{schonberger2016structure,schonberger2016pixelwise}.
The groundtruth poses for the input images were initially calculated by LiDAR SLAM and further refined by bundle adjustment with prior results. 
Afterwards, metric depth images were generated by the COLMAP MVS algorithm. 
For evaluation, we accumulated 0.3 seconds of LiDAR scans corresponding to one scan line since the LiDAR is sampled at 10 Hz. It is difficult to accumulate more LiDAR sweeps because pedestrians cause severe noise and artifacts in the projected depth maps.

\noindent\textbf{TUM RGB-D:}
The TUM RGB-D dataset \cite{sturm2012benchmark} is comprised of 39 video sequences recorded with a Microsoft Kinect. We used a subset of sequence, \textit{walking}, which contains moderate dynamic scenes that two people walk both in the background and the foreground. We train our method on these subsets and evaluate on the test set following MC \cite{li2019learning}.  

\noindent\textbf{NYUv2:}
 Although the NYUv2 dataset \cite{silberman2012indoor} does not contain dynamic scenes, it is the most common public benchmark of depth estimation for indoor scenes. The training dataset consists of 268K images. We subsampled every 10 frames and used only 47k frames for training. The official 654 test scenes are used for evaluation.


\noindent\textbf{Implementation Details:} Our monocular depth estimation network is based on U-Net \cite{ronneberger2015u}, an encoder-decoder architecture. 
 For training, we apply FlowNet2.0 \cite{ilg2017flownet} to compute optical flow and Mask R-CNN \cite{he2017mask} pre-trained on the COCO dataset \cite{lin2014microsoft} to detect human masks. For normal-guided scale constraint, we apply $20 \times 20$ boxes to compose the small patch $B$. The camera of the NAVERLABS dataset is towards the upper side for $\theta = 10^{\circ}$ as in Fig.\:\ref{fig:surface normal} (b); thus $S_{th}$ is set to $15^{\circ}$. We investigate the sampling ratio for $M^{\prime}$ in Eq.\:\ref{eq:weqk_supervision} at 10\%, 30\%, and 50\%, and we select 30\% because it is slightly better than others. In the supplementary material, we provide more implementation details of network architecture and training procedures.     

\subsection{Experimental Results on NAVERLABS dataset}

\begin{figure*}[t]
    \centering
    \includegraphics[width=0.8\linewidth]{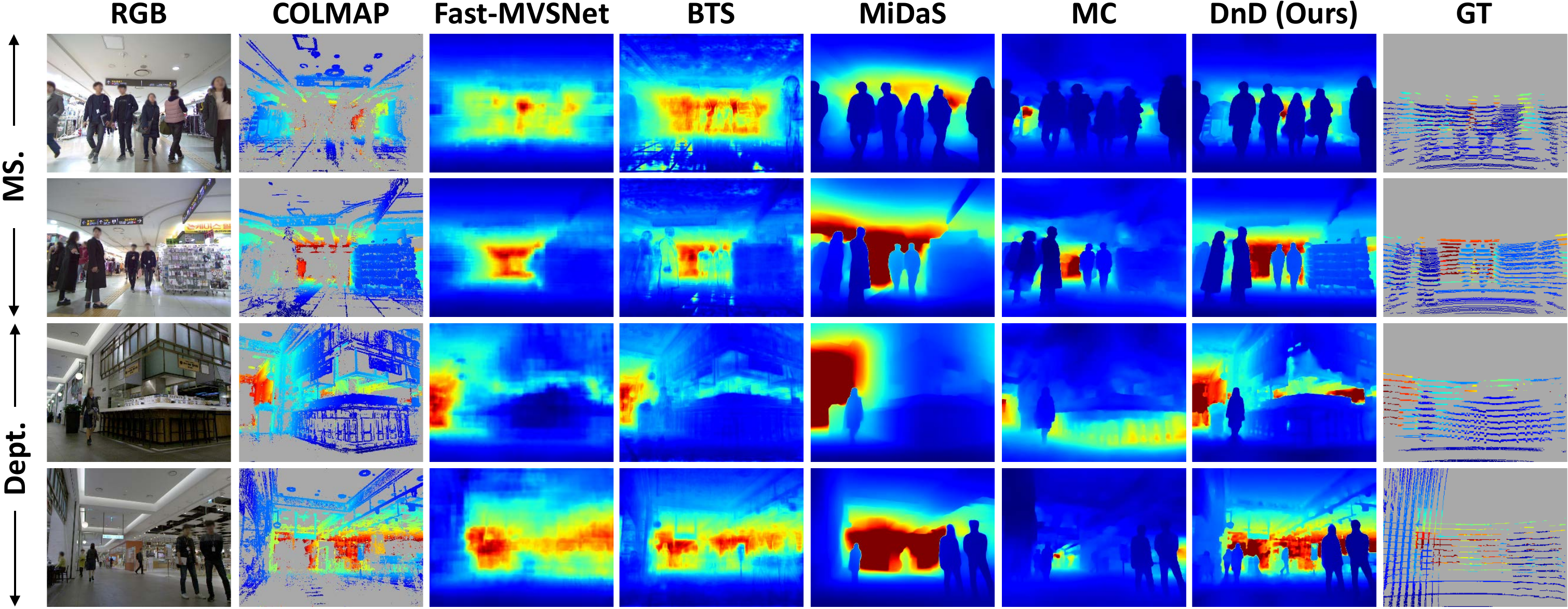}
    \vspace{-2mm}
    \caption{Qualitative comparisons of the Metro Station dataset (1-2 rows) and the Department Store dataset (3-4 rows). Our network takes an image (1st column) and a depth input projected from the 3D model (2nd column), and predicts dense depth maps (7th column). Ground truth (8th column) collected from 3D LiDAR is used to evaluate our method. Depth results (3-6 columns) are estimated by the existing depth estimation algorithms. Compared to MiDaS and MC, DnD generates detailed and high quality depth maps in non-human regions and also shows depth maps in human regions with sharp depth discontinuities. All color maps use the jet color map (low \includegraphics[width=0.5cm,height=0.2cm]{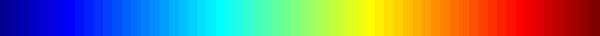} high; grey means empty depth values).  
    }
    \label{fig:qualitative}
\end{figure*}

\begin{table*}[t]
    \centering
    \resizebox{0.9\linewidth}{!}{
    \normalsize\begin{tabular}{l||cccc|ccc}
    \toprule
    \multirow{2}{*}{Method} & \multicolumn{4}{c|}{F+B / F ( Lower is better )} & \multicolumn{3}{c}{F+B / F ( Higher is better )} \\ 
    & Abs Rel & Sq Rel & RMSE & RMSE log & $\delta_{1.25}$ & $\delta_{1.25^2}$ & $\delta_{1.25^3}$\\ \midrule
    DnD ($L_{ph}$\:\text{only}) & 0.326 / 0.596 & 1.38 / 4.16 & 2.39 / 3.60 & 0.107 / 0.177 & 0.711 / 0.545 & 0.840 / 0.705 & 0.898 / 0.795 \\
    DnD w/o FSC,\:NSC & 0.209 / 0.304 & 0.24 / 0.25 & 1.80 / 2.56 & 0.095 / 0.172 & 0.767 / 0.540 & 0.851 / 0.690 & 0.899 / 0.787 \\
    DnD w/o NSC & 0.204 / 0.299 & 0.23 / 0.24 & 1.78 / 2.51 & 0.094 / 0.170 & 0.772 / 0.552 & 0.856 / 0.701 & 0.901 / 0.790\\
    DnD w/o FSC & 0.192 / 0.252 & \textbf{0.20} / 0.17 & 1.77 / 2.47 & 0.086 / 0.139 & 0.794 / 0.644 & 0.877 / 0.785 & 0.917 / 0.851 \\
        DnD (full) & \textbf{0.189} / \textbf{0.240} & \textbf{0.20} / \textbf{0.16} & \textbf{1.76} / \textbf{2.44} & \textbf{0.084} / \textbf{0.133} & \textbf{0.806} / \textbf{0.677} & \textbf{0.881} / \textbf{0.798} & \textbf{0.919} / \textbf{0.856} \\
    \bottomrule
    \end{tabular}
    }
    \vspace{-2mm}
    \caption{Contributions of our proposed modules to the evaluation results on the Metro Station (MS) dataset. F means the evaluation results on depth values in the human regions and F+B indicates the evaluation results on depth values over the entire scene. The median scaling is applied to DnD ($L_{ph}$ only) for absolute scale depth prediction. Compared to DnD ($L_{ph}$ only), we improve the RMSE by 26.4\% in the entire scene and by 32.2\% in the human regions.}
    \vspace{-4mm}
    \label{table:ablation}
\end{table*}

For quantitative evaluation, we compare our proposed method DnD with five different methods: 1) Fast-MVSNet \cite{yu2020fast} is a learning-based multiview stereo algorithm, 2) DepthComple is a learning-based depth completion algorithm that adopts an early-fusion encoder-decoder network from \cite{ma2019self} combined with normalized convolution layers \cite{eldesokey2019confidence}, 3) MiDaS \cite{ranftl2020towards} is a monocular depth estimation method trained on a diverse set of datasets including various dynamic scenes, 4) MC \cite{li2019learning} is a monocular depth estimation method trained on a Mannequin Challenge dataset, and 5) BTS \cite{lee2019big} is the state-of-the-art supervised monocular depth estimation method on the benchmark datasets \cite{silberman2012indoor}. To experiment with Fast-MVSNet, DepthComple, and BTS, we use their public codes and train the network from scratch with the NAVERLABS dataset. MiDaS and MC are implemented with pre-trained weights from their public codes. To train the former methods, we exploit a depth map projected from the 3D model as groundtruth. 
In our experiments, training video-based depth estimation methods \cite{luo2020consistent,Teed2020DeepV2D} fails in the video frames of NAVERLABS. Their methods based on static scene assumption cannot be applied for scenes with extreme object motion. Furthermore, self-supervised methods \cite{godard2019digging,ma2019self} show poor performance for depth evaluation. We observe that a joint training framework of pose and depth from a monocular video is extremely difficult because pose networks often fail to estimate proper camera ego-motion in complex and crowded indoor environments. Instead, we provide experimental results of DnD ($L_{ph}$ only) trained only with the photometric consistency loss similar to self-supervised training settings, e.g., Monodepth2 \cite{godard2019digging}. DnD ($L_{ph}$ only) only converts an RGB image to a dense depth map and thus suffers from the scale ambiguity issue.      
DnD, Fast-MVSNet, DepthComple, and BTS are able to estimate absolute depth values while MiDaS and MC only output relative depth up to an unknown scale factor. Thus, we apply a median scaling \cite{godard2019digging} to define a scale factor by comparing their depth predictions with the absolute depth from our groundtruth depths. In Table \ref{table: metric}, we summarize the quantitative results on two indoor datasets. DnD outperforms other depth estimation methods on all evaluation metrics. In particular, we observe an improvement in performance at depths of both people (F) and the entire scene (F+B). Although the models provided by MiDaS and MC were trained on different datasets, the methods focus on composing datasets for dynamic scenes and training their depth estimation network with supervised learning. Compared to such methods, our method can be scalable to general environments without high quality groundtruth depths. 

A qualitative comparison is shown in Fig.\:\ref{fig:qualitative}. The visual results of DepthComple are added in the supplementary material because this method fails to generate reasonable depth maps. In general, the qualitative comparison corresponds well to our quantitative results in Table \ref{table: metric}. In terms of depth in human regions, Fast-MVSNet and BTS fail to reconstruct the depth for moving people regions. BTS is not able to learn depths for moving people regions because it only relies on COLMAP results as ground truth. Fast-MVSNet is also not able to provide depths for moving people due to the limitation of multi-view matching. In contrast, MC and MiDaS adequately estimate the depths in human regions with sharp depth discontinuities. However, these methods show blurry and noisy depths for background regions. Compared to other methods, depth predictions over background regions from our method are sharper and less noisy.

\begin{figure}
    \centering
    \includegraphics[width=0.9\linewidth]{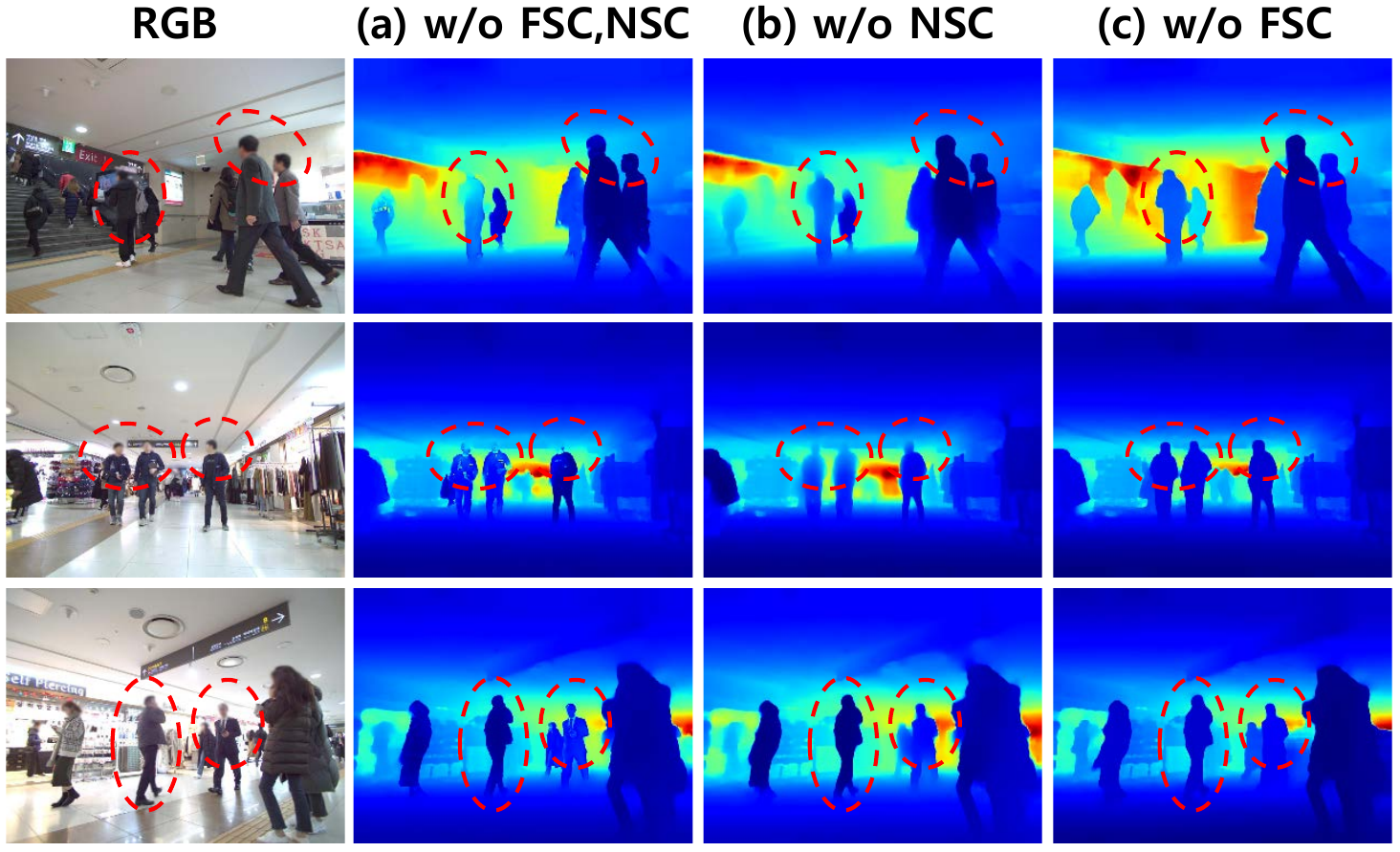}
    \vspace{-3mm}
    \caption{Qualitative comparisons of the proposed modules. (a) We note that a part of the human is cut off in the depth map. The visual artifacts that look similar to RGB image patterns often appear in human depth regions. (b) FSC encourages our network to capture sharper details, fill in missing parts of the human regions, and remove visual artifacts. (c) We observe that NSC estimates depth with sharp depth discontinuities and also recovers absolute scale depth values for human regions. 
    }
    \vspace{-4mm}
    \label{fig:ablation}
\end{figure}

\noindent\textbf{Ablation Study}
In Table \ref{table:ablation}, we validate the influence of different loss terms within our proposed method. We note that all our proposed modules contribute to a significant improvement in Table \ref{table:ablation}. In particular, both NSC and FSC significantly improve depth estimation performance for all evaluation metrics.  Figure \ref{fig:ablation} provides an ablation study using qualitative comparison to better understand the effectiveness of different loss terms in our method. In Fig.\:\ref{fig:ablation} (a), only using depth loss and photometric consistency loss for training is able to estimate dense depth maps over the entire scene. However, they show poor results especially on depth in human regions. Figure\:\ref{fig:ablation} (b) and (c) validate that our method improves the depth estimation performance with each proposed module.

\begin{figure}
    \centering
    \includegraphics[width=\linewidth]{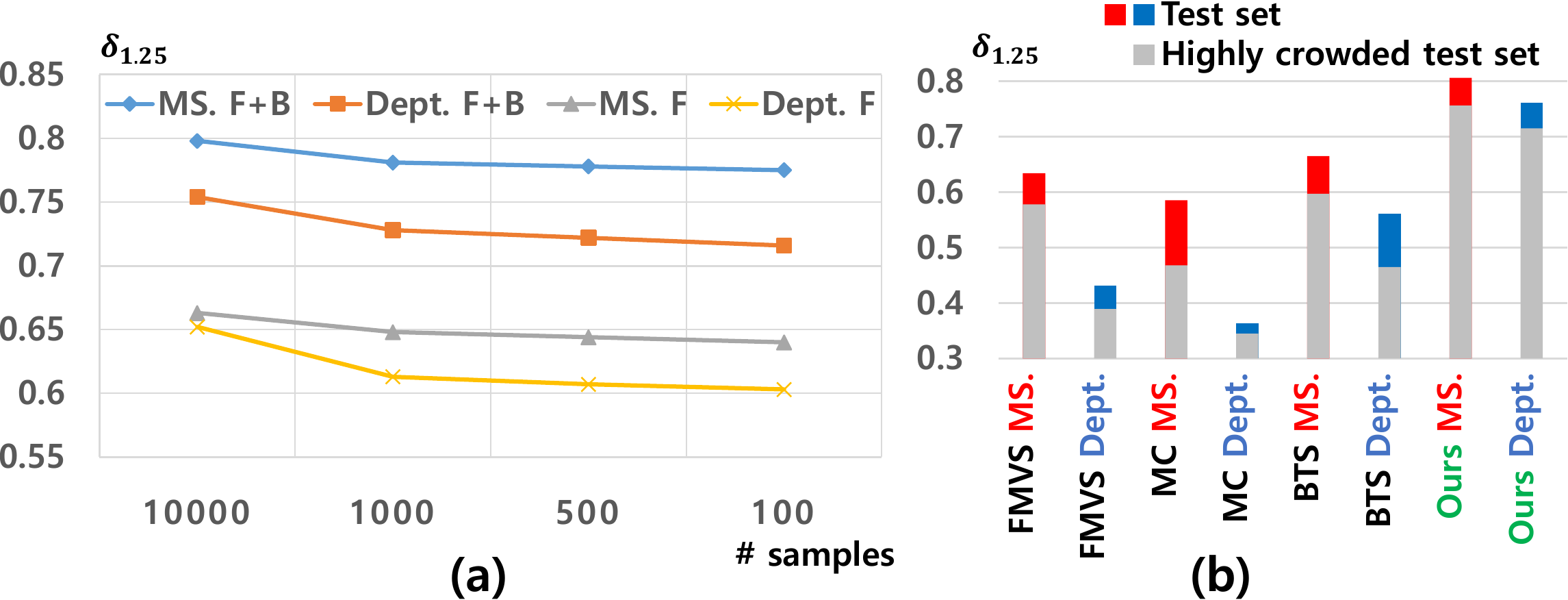}
    \vspace{-6mm}
    \caption{(a) shows $\delta_{1.25}$ metric for our method corresponding the number of uniform-sampled input depth points, and (b) indicates the performance degradation for scenes with high density crowds in MS and Dept: 8.8\% and 9.7\% for FMVS (Fast-MVSNet \cite{yu2020fast}), 20\% and 5.2\% for MC \cite{li2019learning}, 10.2\% and 17.1\% for BTS \cite{lee2019big}, 6.2\% and 6\% for ours.}
    \label{fig:graphl}
\end{figure}

\noindent\textbf{Analysis} We study the effectiveness of the number of sparse depth points and the crowd density. We use the accuracy score $\delta_{1.25}$ as an evaluation metric because other metrics do not show significant performance degradation in DnD.       
If the number of matched local features is not enough for reconstructing a plausible 3D model, the projected depth input for DnD might become a highly sparse map.
In this case, the experimental results in Fig.\:\ref{fig:graphl} (a) show the relationship between the number of depth points and the performance of DnD.
As the number of input depth points decreases, the performance degrades slightly as expected.
However, this robustness to the sparsity proves that the RGB images serve as the primary source of our monocular depth estimation network.
Furthermore, for the dense crowded scenes in indoor environments, we select some highly crowded samples in the original test set for the NAVERLABS dataset. These are composed of 12\% or more human pixels out of the total number of the pixels in one image. 
Note that the crowd density is a more significant factor than the number of people per scene in our task. We focus on pedestrians whose apparent sizes are large (i.e., high crowd density) because they are close to the camera and occlude those who lie behind them.      
Despite the multiple non-rigid people, we notice that DnD robustly estimates depth for both humans and background regions as in Fig.\:\ref{fig:graphl} (b).

\noindent \textbf{DnD with Localization} For the complete system of DnD, there is a need for reprojected depth maps obtained from the 3D model using visual localization techniques. 
Thus, we exploit Kapture toolbox \cite{pion2020benchmarking} to estimate the camera pose of given input images, and then project the 3D model to the sparse depth maps. 
We describe more details about the whole pipeline in the supplementary materials.


\subsection{Results on NYUv2 and TUM RGB-D datasets}
We train and evaluate our method on each NYUv2 and TUM RGB-D dataset. 
Since the 3D map for TUM and NYUv2 produced by COLMAP is only up to an unknown scale factor, we cannot use metric depth as input. Instead, we obtain the metric depth maps by sparsifying depth maps collected from Kinect in order to use standard error metrics for a fair comparison. To simulate sparse patterns existing in COLMAP, we only maintain the depth values according to the location of SIFT features \cite{Lowe2004} in the corresponding images. Since the number of SIFT features varies in each frame, we set the maximum number of features to 200 samples.
Table \ref{table:NYUv2} reports the quantitative comparisons with several other depth estimation methods on both datasets. DnD outperforms other single view methods, demonstrating the benefit of our method in less crowded benchmark indoor datasets.     
In supplementary materials, we show experimental results on TUM when we consider normalized relative depth as projected depth maps obtained from a scale ambiguous 3D model.


\begin{table}[t]
\centering
\resizebox{0.48\textwidth}{!}{
\begin{tabular}{l|c|cc|ccc}
\toprule
\multirow{2}{*}{Method} & \multirow{2}{*}{Dataset} &  \multicolumn{2}{c|}{Lower is better} & \multicolumn{3}{c}{Higher is better} \\
& & RMSE & Abs Rel & $\delta_{1.25}$ & $\delta_{1.25^2}$ & $\delta_{1.25^3}$\\ 
\midrule
BTS \cite{lee2019big} (Single view) & NYUv2  &  0.392 & 0.110 & 0.885 & 0.978 & 0.994 \\ 
Yang et al. \cite{yang2019dense} (Single view) & NYUv2 &  0.569 & 0.171  & - & - & - \\
MonoDepth2 \cite{godard2019digging} (Single view) & NYUv2 &  0.617 & 0.170  & 0.748 & 0.942 & 0.986 \\
P2Net \cite{yu2020p} (Multi view) & NYUv2 &  0.533 & 0.147  & 0.801 & 0.951 & 0.987 \\
DeepV2D \cite{Teed2020DeepV2D} (Multi view) & NYUv2 &  0.403 & \textbf{0.061}  & \textbf{0.956} & \textbf{0.988} & 0.996 \\
\midrule
DnD (Single view) & NYUv2 & \textbf{0.362} & 0.098 & 0.910 & \textbf{0.988} & \textbf{0.998} \\
\midrule
MC \cite{li2019learning} (Single view) & TUM & 0.840 & 0.204 & 0.664 & 0.931 & 0.981 \\
MC \cite{li2019learning} (Multi view) & TUM & \textbf{0.570} & \textbf{0.129} & - & - & - \\
MiDaS \cite{ranftl2020towards} (Single view) & TUM & 0.819 & 0.196 & 0.678 & 0.911 & 0.965 \\ 
\midrule
DnD (Single view) & TUM & 0.631 & 0.175 & \textbf{0.751} & \textbf{0.947} & \textbf{0.982} \\
\bottomrule
\end{tabular}
\vspace{-4mm}
}
\caption{\label{table:NYUv2} Quantitative comparisons of the NYUv2 and TUM RGB-D dataset. In both static and dynamic scenes, DnD shows improved performance compared to recent methods.}
\vspace{-3mm}
\end{table}

\section{Conclusion, Limitations, and Future Work}
In this paper, we present a learning-based method for estimating dense depth maps of dynamic scenes collected from a moving monocular camera. Given RGB images and sparse depth maps projected from a 3D model, our method is able to predict absolute scale depth for multiple pedestrians and complex backgrounds. In complex and crowded indoor environments, this is a practical method to use for various robotics and augmented reality applications. Once we initially have the 3D model, our method enables any freely moving mobile devices with a single-view camera to estimate dense depth maps.
Our method still has limitations that we plan to address.  
The sparse depth maps projected from the 3D model may be inaccurate or not aligned with the current image. Also, our training method relies on FlowNet2.0 \cite{ilg2017flownet} and Mask R-CNN \cite{he2017mask}. 
In the future, we would like to extend our approach to outdoor scenes.

\vskip 1mm
\noindent \textbf{Acknowledgements} This work was supported by the Institute of Information \& communications Technology Planning \& Evaluation(IITP) grant funded by the Korea government(MSIT) (No. 2019-0-01309, Development of AI Technology for Guidance of a Mobile Robot to its Goal with Uncertain Maps in Indoor/Outdoor Environments)

{\small
\bibliographystyle{ieee_fullname}
\bibliography{main}
}

\end{document}


\title{DnD: Dense Depth Estimation in Crowded Indoor Dynamic Scenes Supplementary Material}

\author{Dongki Jung$^{\thanks{These two authors contributed equally.}}$\:\:\textsuperscript{\rm 1} \:\: Jaehoon Choi$^{\footnotemark[1]}$\:\:\textsuperscript{\rm 1,2} \:\: Yonghan Lee$^{1}$ \:\: Deokhwa Kim$^{1}$ \:\: Changick Kim$^{3}$ \\ 
Dinesh Manocha$^{2}$ \:\: Donghwan Lee$^{1}$ \\ 
$^{1}$NAVER LABS \:\: $^{2}$University of Maryland \:\: $^{3}$KAIST 
}

\maketitle
\ificcvfinal\thispagestyle{empty}\fi
\appendix




\section{Datasets}
 We provide a brief overview of the NAVERLABS Indoor Localization dataset \cite{SpoxelNet,NAVIdataset} and the specific properties that make it desirable as a dense depth estimation in crowded indoor dynamic scenes. The NAVERLABS dataset consists of various video sequences captured on five different floors in a department store and a metro station. Among these places, we selected two places, the metro station B1 (MS) and the department store B1 (Dept), to build a benchmark for depth estimation in crowded indoor dynamic scenes. 
 The MS dataset is one of the most crowded dataset in the NAVERLABS dataset.
 We also chose the Dept dataset because it has totally different environments. In the NAVERLABS dataset, a mapping robot utilizes two 16-channel LiDAR sensors, six RGB cameras and four smartphone cameras. We only utilized images collected from six RGB cameras. LiDAR sensors are only used for generating groundtruth depth maps for evaluation. 
As the NAVERLABS dataset consists of the split of 60735 images for training and 835 images for testing in MS, and 25083 images for training and 443 images for testing in Dept, we adopt this dataset for dense depth estimation in crowded indoor dynamic scenes.   

\begin{table}[h]
    \centering
    \resizebox{\linewidth}{!}{\normalsize\begin{tabular}{lccccccc}
    \toprule
    Dataset & Environment & Location & \textcolor{red}{\textbf{Dynamic}} & Video & \textcolor{red}{\textbf{\# Images}} & \textcolor{red}{\textbf{Metric}} & Annotation  \\ 
    \midrule
    (a) Indoor datasets\\ 
    NYUv2 \cite{silberman2012indoor} & Indoor & Small Rooms  & & \checkmark & 407K & \checkmark & RGB-D \\
    ScanNet \cite{dai2017scannet} & Indoor & Small Rooms & & \checkmark & 2.5M & \checkmark & RGB-D \\
    Stanford \cite{armeni2017joint} & Indoor & Small Rooms & & \checkmark & 72K & \checkmark & Laser \\
    Matterport3D \cite{chang2017matterport3d} & Indoor & Small Rooms & & \checkmark & 194K & \checkmark & Laser \\
    7 scenes \cite{shotton2013scene} & Indoor & Small Rooms & & \checkmark & 26K & \checkmark & Laser \\
    InLoc \cite{taira2018inloc} & Indoor & Univ. bldg. & & \checkmark & 10K & \checkmark & Laser \\
    Baidu \cite{sun2017dataset} & Indoor & Mall & \checkmark & \checkmark & 682 & \checkmark & Laser \\
    \midrule
    (b) Internet Image collections \\ 
    DIW \cite{chen2016single} & Indoor \& Outdoor & Diverse & \checkmark & & 495K & & Ordinal  \\
    MegaDepth \cite{li2018megadepth} & Outdoor & Diverse &  & & 130K & & SfM \\
    ReDWeb \cite{xian2018monocular} & Indoor \& Outdoor & Diverse & \checkmark & \checkmark & 3600 & & Stereo \\
    WSVD \cite{wang2019web} & Indoor \& Outdoor & Diverse & \checkmark & \checkmark & 150K & & Stereo \\
    MC \cite{li2019learning} & Indoor \& Outdoor & Diverse & \checkmark & \checkmark & 136K & & \\
    \midrule
    (c) \textbf{NAVERLABS}\\ 
    \textbf{Dept} \cite{SpoxelNet,NAVIdataset} & Indoor & Mall  & \checkmark & \checkmark & 25K & \checkmark & Laser  \\
    \textbf{MS} \cite{SpoxelNet,NAVIdataset} & Indoor & Mall \& Turnstiles & \checkmark & \checkmark & 60K & \checkmark & Laser  \\
    \bottomrule
    \end{tabular}}
    \caption{(a) Most large-scale indoor datasets do not have dynamic scenes except for Baidu \cite{sun2017dataset}. (b) All datasets collected from the internet do not contain metric 3D models. To train depth estimation algorithms in crowded indoor dynamic scenes, a dataset must contain three properties: dynamic, metric and large amount of images. Different from previous datasets (a) and (b), NAVERLABS contains large amounts of dynamic scenes with the metric 3D model in indoor environments.}
    \label{table:dataset_comparison}
\end{table}
Dense depth estimation in crowded indoor dynamic scenes is a significant task for Robotics and AR applications. To the best of our knowledge, however, it is difficult to find large-scale public datasets for dense depth estimation in crowded indoor dynamic scenes. 
In Table \ref{table:dataset_comparison} (a), most indoor datasets \cite{silberman2012indoor,dai2017scannet,armeni2017joint,chang2017matterport3d,shotton2013scene,taira2018inloc,sun2017dataset} are captured from small collections of room, office, and university buildings. They cover only a restricted scale of spaces and have similar design and internal structures. Additionally, existing indoor datasets do not provide scenes containing multiple dynamic objects such as moving people except for the Baidu dataset \cite{sun2017dataset}. The Baidu dataset \cite{sun2017dataset} collected from a shopping mall has dynamic scenes with 3.75\% crowd density. However, Baidu contains only 689 images for training. In Table \ref{table:dataset_comparison} (b),  other works \cite{chen2016single,li2018megadepth,xian2018monocular,wang2019web,li2019learning} explore the use of internet photo collections to train the depth estimation models for dynamic scenes. These datasets contain a large number of dynamic scenes, but they do not provide metric depth. They often have relative depth obtained by either stereo matching or COLMAP \cite{schonberger2016structure,schonberger2016pixelwise}, or ordinal depth relation from manual annotations. Thus, we are not able to evaluate our algorithms using metric depth maps for internet images.     

\begin{figure}[b]
    \centering
    \includegraphics[width=0.9\linewidth]{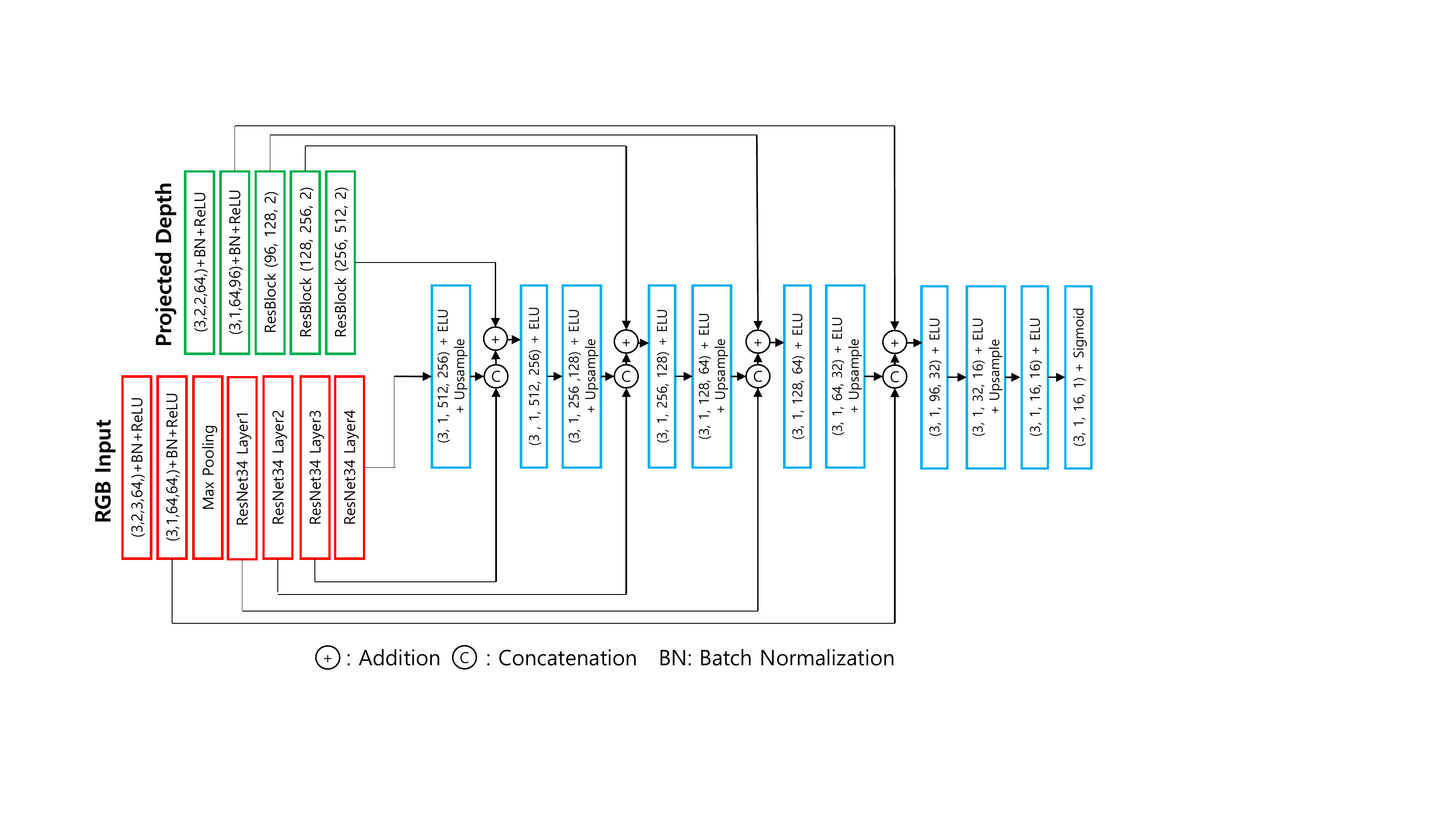}
    \caption{The network architecture of our proposed method. The red block, green block, and blue block are an RGB encoder, a depth encoder, and a decoder.
    We concatenate a projected depth map and a binary mask, which indicates where the projected depth values exist.
    The convolutional filter is defined as (filter size, stride size, input channel, output channel). The RGB encoder adopts a ResNet 34 \cite{he2016deep} backbone pretrained with ImageNet \cite{russakovsky2015imagenet}, and the depth encoder consists of two convolutional blocks and three residual blocks.}
    \label{fig:NetworkDetails}
\end{figure}

\section{Evaluation Metrics}
The depth value is represented in the metric scale (m). Following \cite{eigen2014depth}, we evaluate our method using the following metrics: absolute relative difference (Abs Rel), square relative difference (Sq Rel), root mean squared error (RMSE), RMSE in logarithmic scale (RMSE log), and $\delta_{i}$ meaning the percentage of predicted pixels for which the relative error is less than a threshold \(i\).

\section{Implementation Details}
We implement our method in PyTorch \cite{paszke2017automatic} and conduct all experiments on a V100 GPU. We train our network with a batch size of 8 images with size  1024\(\times\)768 for 20 epochs. We use the following set of weights for each loss term in the loss function: \(\lambda_{d}\) = 0.001, \(\lambda_{ph}\) = 1, \(\lambda_{s}\) = 0.3, \(\lambda_{f}\) = 0.1, and \(\lambda_{n}\) = 0.001. We utilize the Adam \cite{kingma2014adam} with \(\beta_{1}=0.9\) and \(\beta_{2}=0.999\). We use an initial learning rate of \(10^{-4}\) for the first 10 epochs and halve it for the remaining 10 epochs. We use human masks to eliminate the points in human regions to reduce the noise on projected depth maps. 


\section{DnD with Localization}
\begin{figure}[h]
    \centering
    \includegraphics[width=0.8\linewidth]{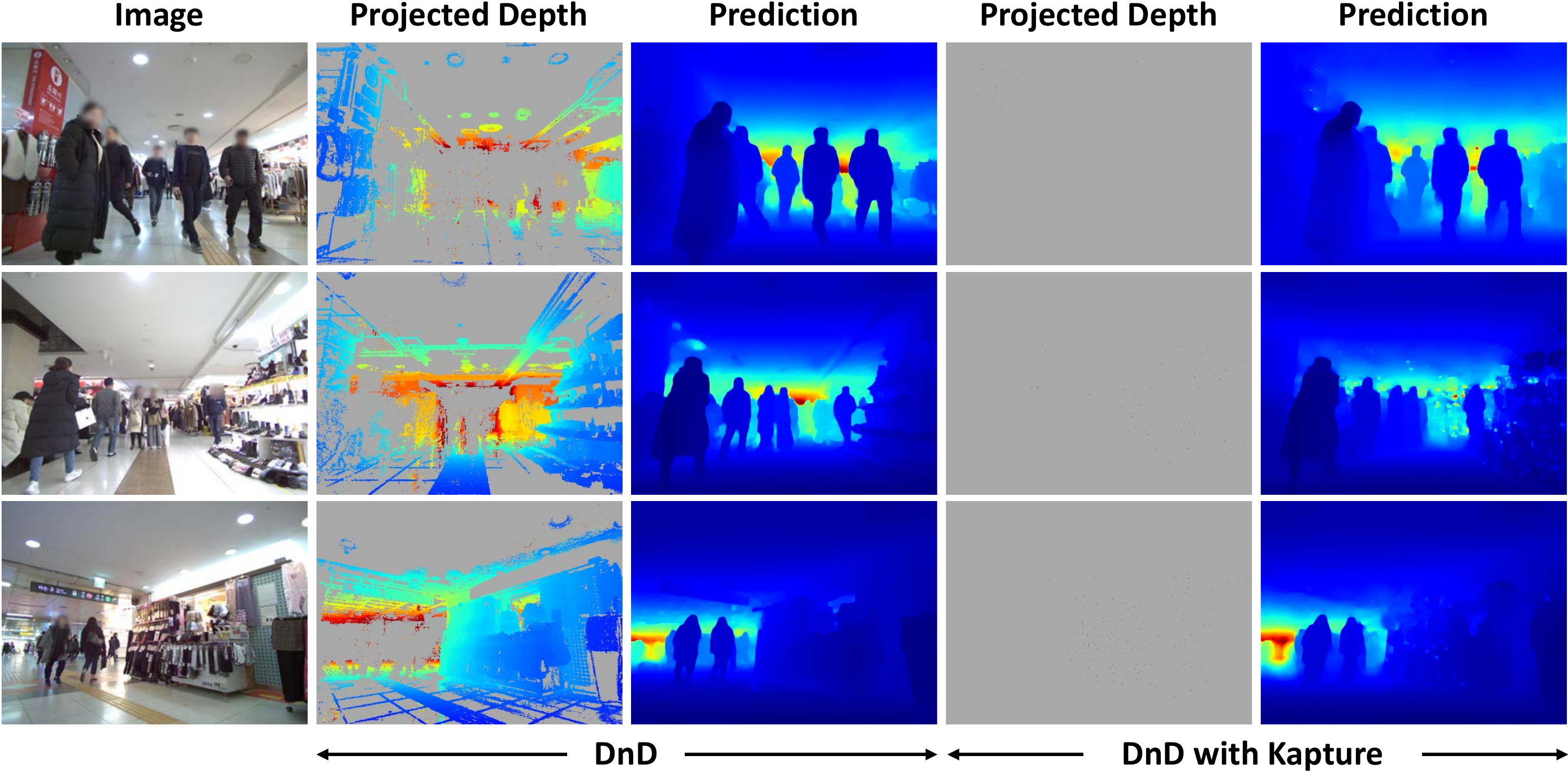}
    \vspace{-3mm}
    \caption{DnD (2-3 columns) estimates depth from the image and the projected depth map, which is derived from the 3D model. Since DnD with Kapture \cite{benchmarking_ir3DV2020} (4-5 columns) exploit the sparse 3D model from SfM, the projected depth maps are also highly sparse, and the background details of the estimated depth maps decrease. However, the overall qualities, especially moving people, are predicted robustly. (low \includegraphics[width=0.5cm,height=0.2cm]{figures/colorbar.jpeg} high; grey means empty depth values.)}
    \label{fig:VL}
\end{figure}

DnD network requires a monocular image and the corresponding reprojected depth map derived from the 3D model with the camera position.
In order to verify the applicability, it is necessary to demonstrate both pose estimation and depth map projection about the new test images.
Thus, we implement the whole pipeline of DnD, including mapping and localization process using Kapture \cite{benchmarking_ir3DV2020}, which is the visual localization toolbox.
As a mapping procedure, the sparse 3D model is reconstructed via Structure-from-Motion (SfM) \cite{schonberger2016structure} in the set of training images.
When a test image is given, image retrieval (AP-GeM \cite{revaud2019learning}) is performed to obtain top $k$ ranked images in the database of the 3D model, which can consider covisibility with the images and estimate a more accurate camera pose.
The local feature extractor (R2D2 \cite{r2d2}) produces 2D-2D matches between the test image and top-ranked database images, resulting in 2D-3D matches between the test image and the 3D model.
Perspective-n-Point (PnP \cite{lepetit2009epnp}) problem with Random Sample Consensus (RANSAC \cite{fischler1981random}) is solved in these matching points and consequently yields the predicted camera pose.
With the visible 3D points from the test image and the estimated pose, we can obtain the reprojected depth map and activate our DnD framework.
We assign $5$ for the top $k$, and the results of the MS dataset are reported in Fig. \ref{fig:VL} and Table \ref{table:kapture_localization}.

\begin{table*}[h]
    \centering
    \resizebox{0.9\textwidth}{!}{\normalsize
    \begin{tabular}{l|c|cccc|ccc}
    \toprule
    \multirow{2}{*}{Method} & \multirow{2}{*}{3D Model} & \multicolumn{4}{c|}{F+B / F ( Lower is better )} & \multicolumn{3}{c}{F+B / F ( Higher is better )} \\
    & & Abs Rel & Sq Rel\: & RMSE & RMSE log & $\delta_{1.25}$ & $\delta_{1.25^2}$ & $\delta_{1.25^3}$\\ \hline
    DnD & SfM + MVS & 0.189 / 0.240 & 0.20 / 0.16 & 1.76 / 2.44 & 0.084 / 0.133 & 0.806 / 0.677 & 0.881 / 0.798 & 0.919 / 0.856 \\
    \midrule
    DnD & SfM & 0.194 / 0.249 & 0.17 / 0.15 & 1.85 / 2.56 & 0.092 / 0.141 & 0.774 / 0.638 & 0.874 / 0.785 & 0.918 / 0.850 \\
    \bottomrule
    \end{tabular}
    }
    \vspace{-2mm}
    \caption{The performance with or without Kapture \cite{benchmarking_ir3DV2020} in the MS dataset.
    The upper line in the table shows the results of the camera poses, which are used in reconstruction of the 3D model (SfM + MVS).
    The lower line in the table indicates the performance of the whole pipeline including visual localization under the sparse 3D model from SfM (without MVS). Although the SfM point clouds are highly sparse and visual localization methods has pose uncertainty, the comparable results show the robustness of our framework.}
    \label{table:kapture_localization}
\end{table*}
\vspace{-4mm}

\vspace{-2mm}

\section{Robustness to Orientation Errors}
We assume a scenario that the camera pose is inaccurate and projected depth maps are not aligned well with the current image. This scenario is common in real-world applications because the visual localization system often computes uncertain poses, especially in dynamic scenes where the moving people cover a large part of the image. Therefore, we validate that our method is robust to orientation errors caused by visual localization algorithms. We add noises $\alpha$ to ground truth poses and then project depth maps from the incomplete 3D model by using noisy poses. $\alpha$ is set to $2^{\circ}$, $5^{\circ}$, and $10^{\circ}$. Following the \cite{sarlin2019coarse}, we report the quantitative evaluation with different noises for pose values. We perform this experiment on the MS dataset because it is highly crowded than the Dept dataset.

\begin{table*}[h]
    \centering
    \resizebox{0.9\textwidth}{!}{\normalsize
    \begin{tabular}{l|c|cccc|ccc}
    \toprule
    \multirow{2}{*}{Method} & \multirow{2}{*}{Orientation Error} & \multicolumn{4}{c|}{F+B / F ( Lower is better )} & \multicolumn{3}{c}{F+B / F ( Higher is better )} \\
    & & Abs Rel & Sq Rel\: & RMSE & RMSE log & $\delta_{1.25}$ & $\delta_{1.25^2}$ & $\delta_{1.25^3}$\\ \hline
    DnD & - & 0.189 / 0.240 & 0.20 / 0.16 & 1.76 / 2.44 & 0.084 / 0.133 & 0.806 / 0.677 & 0.881 / 0.798 & 0.919 / 0.856 \\
    \midrule
     & $2^{\circ}$ & 0.205 / 0.246 & 0.24 / 0.17 & 1.82 / 2.46 & 0.089 / 0.135 & 0.786 / 0.666 & 0.876 / 0.796 & 0.917 / 0.855 \\
    DnD & $5^{\circ}$ &  0.207 / 0.246 & 0.23 / 0.17 & 1.83 / 2.46 & 0.090 / 0.135 & 0.779 / 0.666 & 0.875 / 0.796 & 0.917 / 0.855 \\
     & $10^{\circ}$ & 0.211 / 0.245 & 0.24 / 0.17 & 1.84 / 2.46 & 0.091 / 0.135 & 0.773 / 0.666 & 0.873 / 0.795 & 0.916 / 0.855 \\
    \bottomrule
    \end{tabular}
    }
    \label{table:VL}
    \vspace{-2mm}
    \caption{Experimental evaluations of our proposed method's robustness to orientation errors. 
    }
\end{table*}
\vspace{-1mm}

\section{Qualitative Results of Depth Prediction Corresponding Sparse Depth Points}
\vspace{-4mm}
\begin{figure*}[h]
    \centering
    \includegraphics[width=0.56\linewidth]{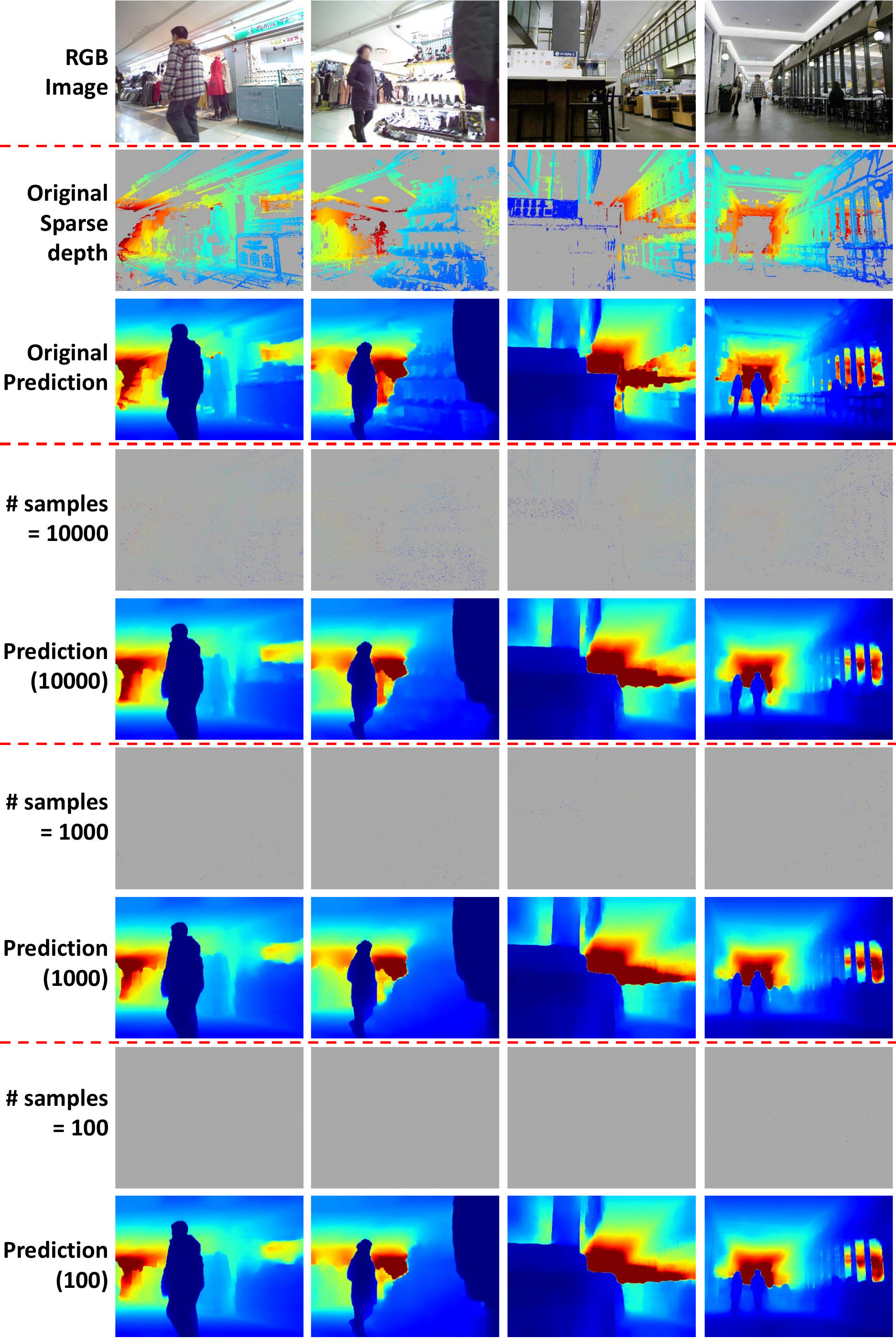}
    \vspace{-2mm}
    \caption{Qualitative results of depth prediction corresponding to the number of uniform-sampled input depth points (refer the Fig. 6 (a) in the manuscript). 
    The first and the second columns show the results of the MS dataset, and the third and the fourth columns indicate the Dept dataset. 
    As the number of sampled points decreases, some details of the static background disappear. However, the overall quality of depth estimation results is maintained robustly. All color maps use the jet color map (low \includegraphics[width=0.5cm,height=0.2cm]{figures/colorbar.jpeg} high; grey means empty depth values.)}
    \label{fig:sparse}
\end{figure*}

\section{Ablation Study}
\begin{table*}[h]
    \centering
    \resizebox{0.9\linewidth}{!}{\normalsize\begin{tabular}{l||cccc|ccc}
    \toprule
    \multirow{2}{*}{Method} & \multicolumn{4}{c|}{F+B / F ( Lower is better )} & \multicolumn{3}{c}{F+B / F ( Higher is better )} \\ 
    & Abs Rel & Sq Rel\: & RMSE & RMSE log & $\delta_{1.25}$ & $\delta_{1.25^2}$ & $\delta_{1.25^3}$\\ \midrule
    DnD (L_{ph}\:\text{only}) & 0.289 / 0.388 & 0.47 / 0.60 & 2.60 / 3.24 & 0.109 / 0.148 & 0.663 / 0.564 & 0.837 / 0.759 & 0.911 / 0.847 \\
    DnD w/o FSC,\:NSC & 0.240 / 0.355 & 0.42 / 0.71 & 2.51 / 3.63 & 0.091 / 0.133 & 0.741 / 0.642 & 0.877 / 0.806 & 0.929 / 0.868 \\
    DnD w/o NSC & 0.226 / 0.335 & 0.37 / 0.72 & 2.39 / 3.19 & 0.087 / 0.130 & 0.753 / 0.635 & 0.885 / 0.810 & 0.930 / 0.878 \\
    DnD w/o FSC & 0.230 / 0.272 & \textbf{0.31} / 0.31 & 2.44 / 3.15 & 0.092 / 0.127 & 0.730 / 0.644 & 0.871 / 0.814 & 0.927 / 0.880 \\
    DnD (full) & \textbf{0.213} / \textbf{0.250} & 0.32 / \textbf{0.30} & \textbf{2.36} / \textbf{3.04} & \textbf{0.084} / \textbf{0.116} & \textbf{0.761} / \textbf{0.707} & \textbf{0.889} / \textbf{0.836} & \textbf{0.932} / \textbf{0.886} \\
    \bottomrule
    \end{tabular}}
    \vspace{-2mm}
    \caption{Contributions of our proposed modules to the evaluation results on the Dept dataset. F means the evaluation results on depth values in the human regions, and F+B indicates the evaluation results on depth values over the entire scene. The median scaling is applied to DnD ($L_{ph}$ only) for absolute scale depth prediction. 
    }
    \label{table:ablation}
\end{table*}

\section{Qualitative Results of Depth Completion Algorithm}
\begin{figure}[h]
    \centering
    \includegraphics[width=0.92\linewidth]{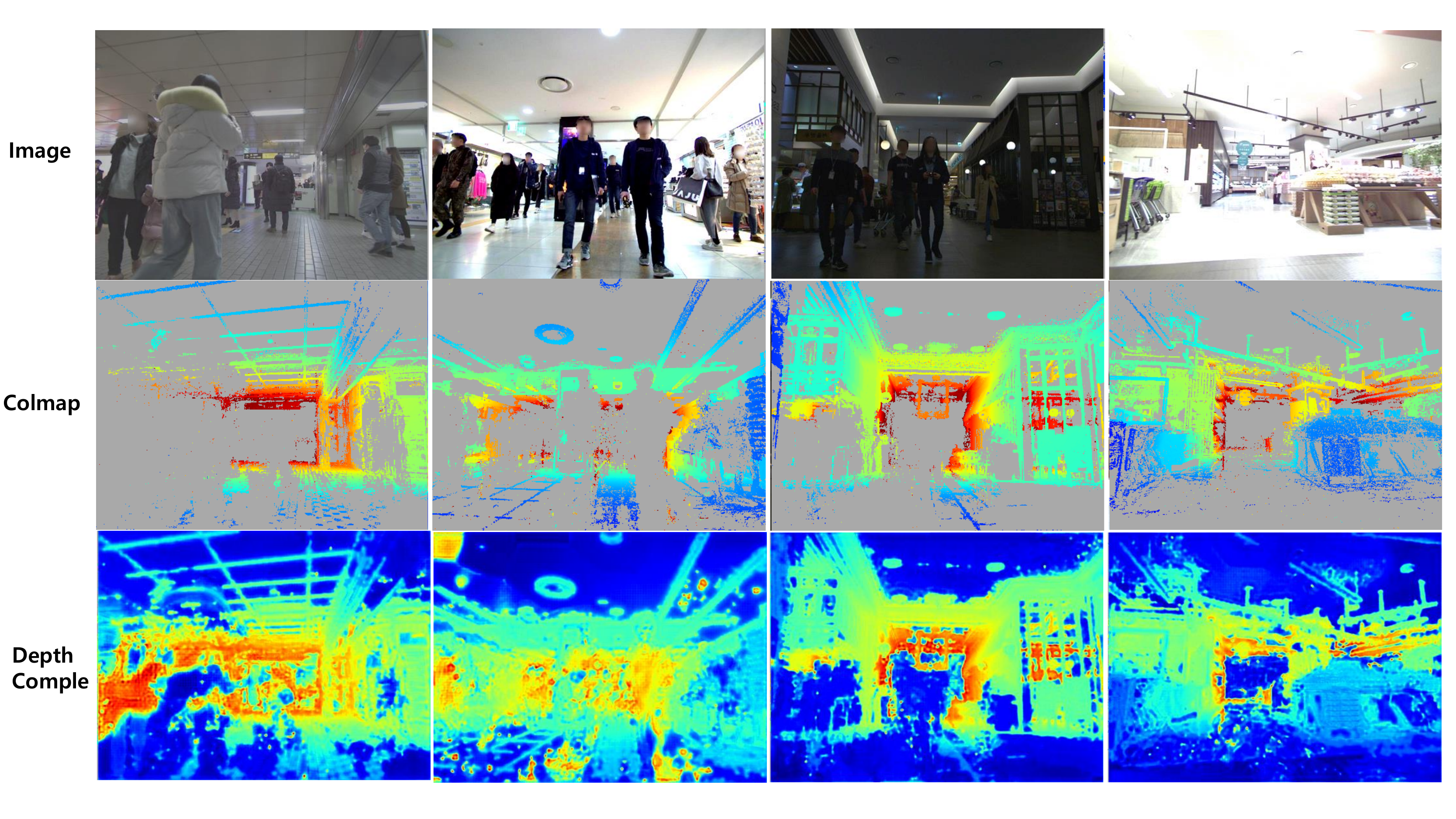}
    \vspace{-4mm}
    \caption{As we metioned in the manuscript, we show the qualitative results of DepthComple \cite{ma2019self}. We adopt an early-fusion encoder-decoder network from \cite{ma2019self} combined with normalized convolution layers \cite{eldesokey2019confidence}. We use projected depth maps as ground truth for supervised training. The depth completion network fails to fill the empty regions in COLMAP results. Also, it fails to produce estimated depth in human regions with accurate depth values. Since most depth completion models show poor performance, we did not add these experimental results in the manuscript. All color maps use the jet color map (low \includegraphics[width=0.5cm,height=0.2cm]{figures/colorbar.jpeg} high; grey means empty depth values.)  
    }
    \label{fig:depthcomple}
\end{figure}
\newpage

\section{Visualization of Intermediate Results for Our Two Novel Consraints}
\begin{figure*}[h]
    \centering
    \includegraphics[width=0.7\linewidth]{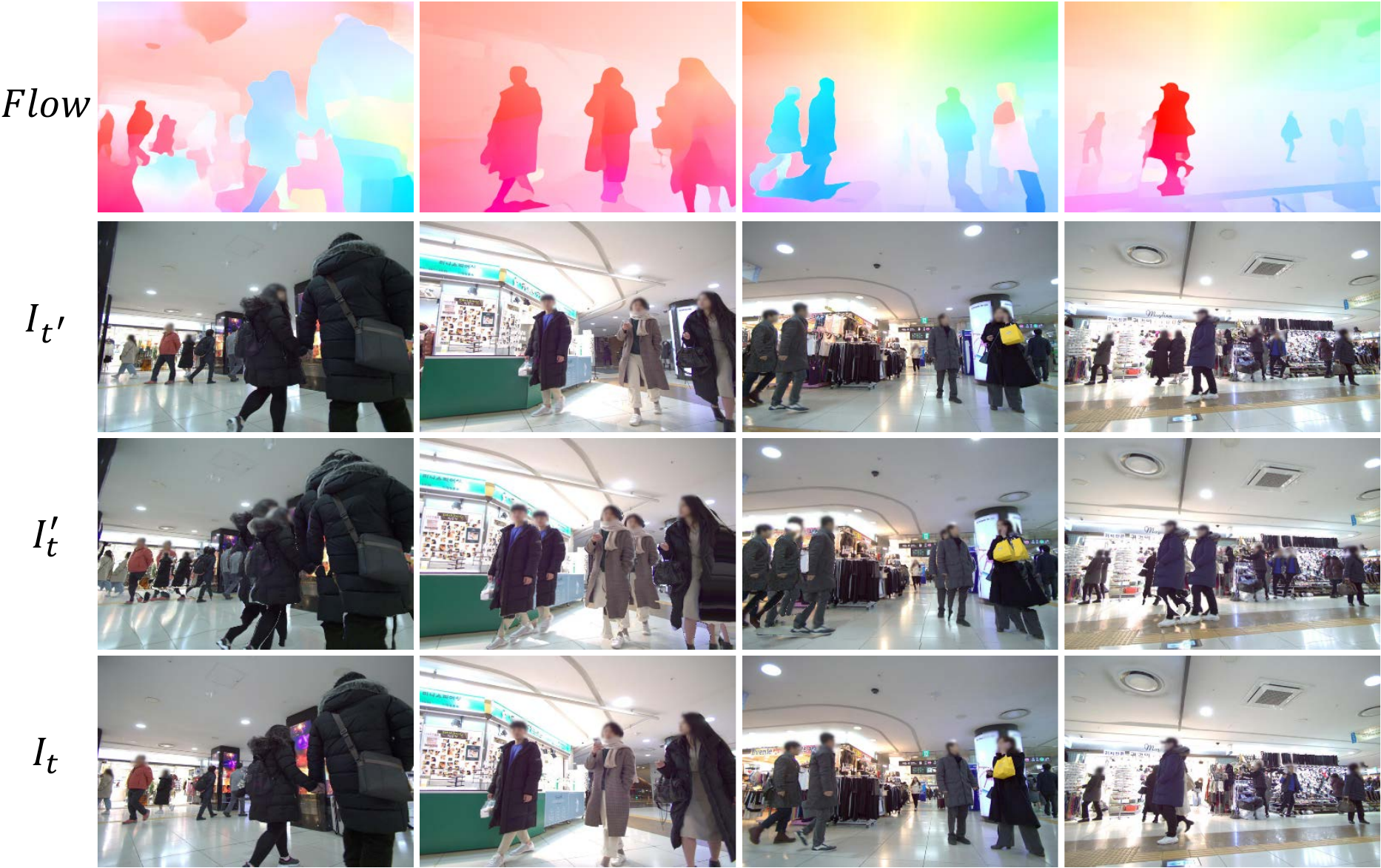}
    \vspace{-2mm}
    \caption{Examples of the optical flow in the MS dataset. From top to bottom, each row shows the optical flows estimated by FlowNet2.0 \cite{ilg2017flownet}, the temporally adjacent images $I_{t^{\prime}}$, the warped images $I^{\prime}_{t}$, and the current images $I_{t}$. 
    Both $I_{t}$ and $I_{t^{\prime}}$ are used as the input for our proposed training method, and the warped images $I^{\prime}_{t}$ are projected from the image coordinates of $I_{t^{\prime}}$ to $I_{t}$ for flow-guided shape constraint (for more details, see Section 3.3 in the manuscript).}
    \label{fig:flow}
\end{figure*}
\vspace{-3mm}

\begin{figure*}[h]
    \centering
    \includegraphics[width=0.7\linewidth]{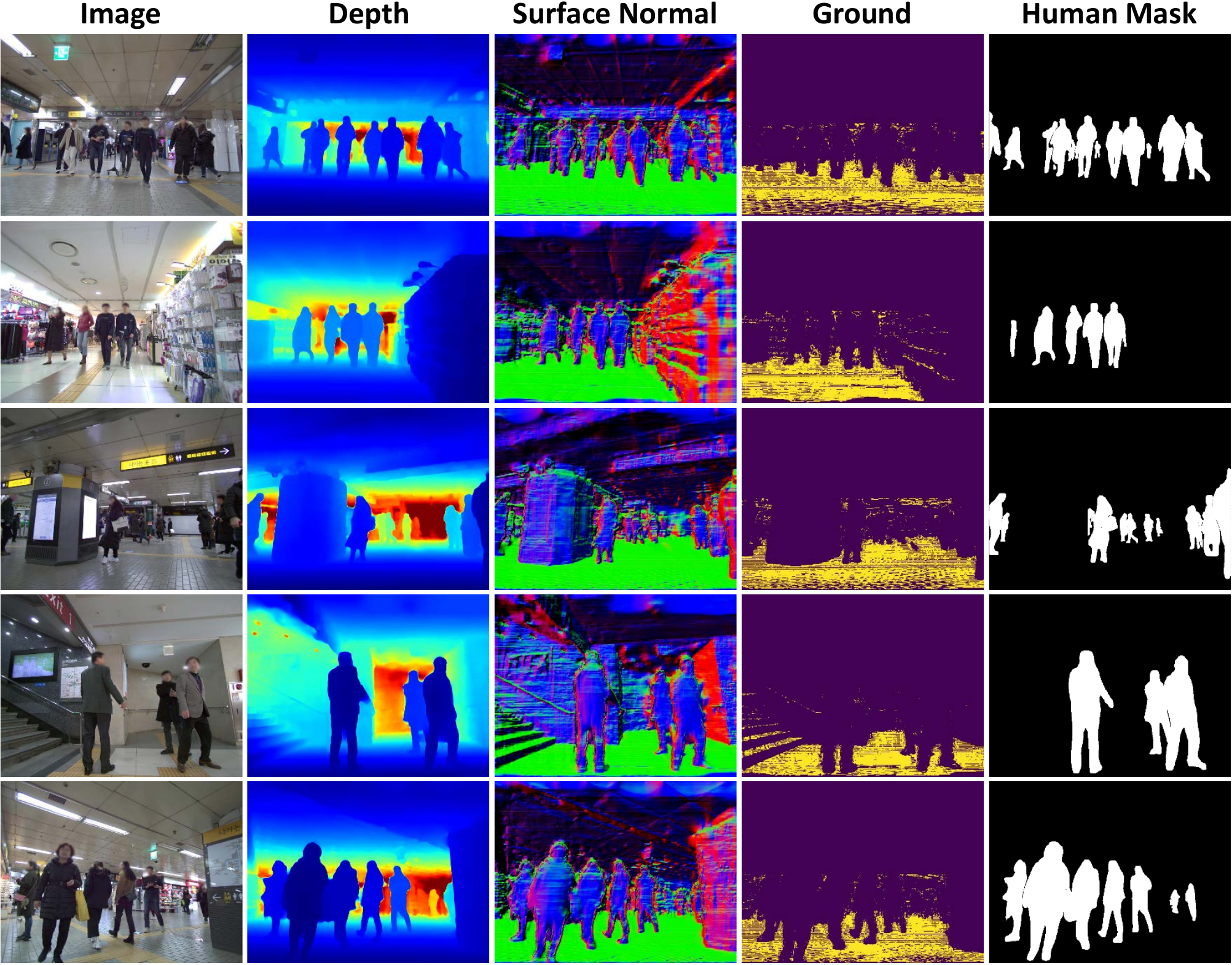}
    \vspace{-2mm}
    \caption{The intermediate results for the normal-guided scale constraints (for more details, see Section 3.4 in the manuscript). We can obtain the normal directions from the estimated depth maps and find the ground regions. Human masks are estimated by Mask R-CNN \cite{he2017mask}. In training, the sampled pixels in each of the people are constrained by the estimated depth values at the human's ground contact point. All color maps use the jet color map (low \includegraphics[width=0.5cm,height=0.2cm]{figures/colorbar.jpeg} high; grey means empty depth values.)}
    \label{fig:normal}
\end{figure*}
\clearpage
\newpage

\section{Results on NYUv2 and TUM RGB-D}
\begin{figure}[h]
    \centering
    \includegraphics[width=0.68\linewidth]{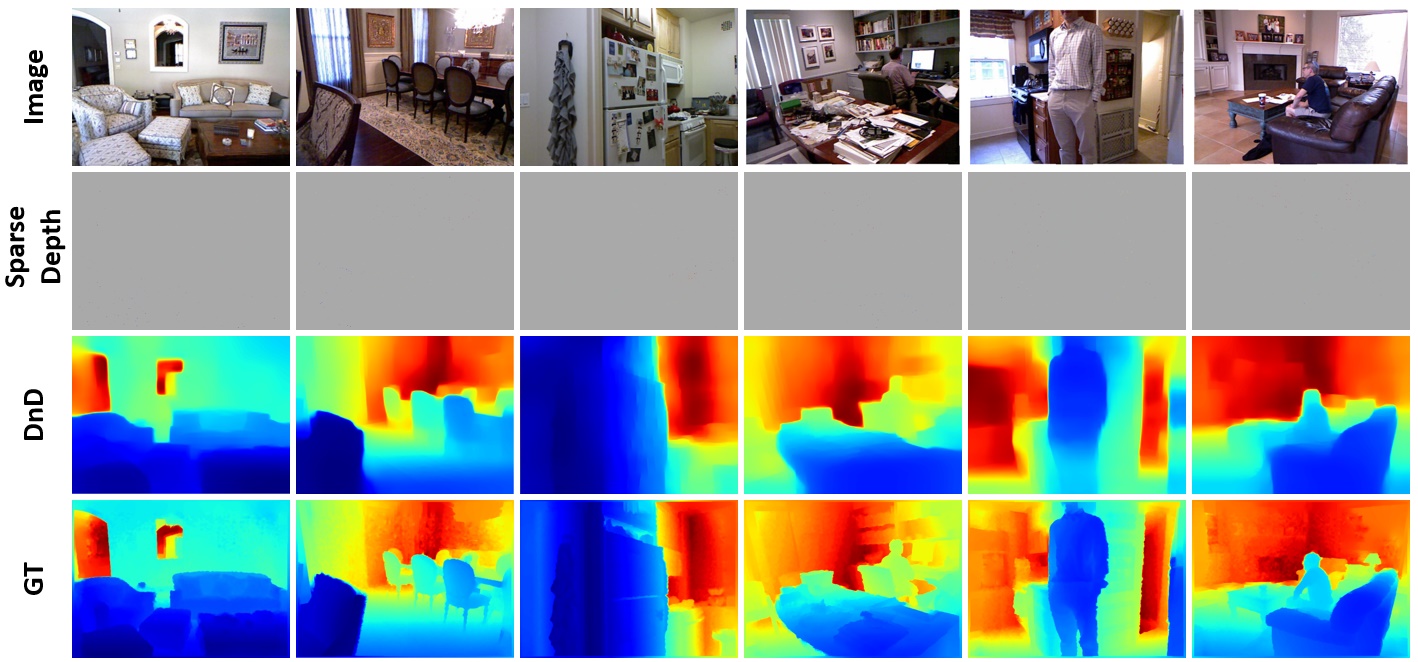}
    \vspace{-2mm}
    \caption{The qualitative results of the NYUv2 dataset.}
    \label{fig:nyu}
    \vspace{-7mm}
\end{figure}

\begin{figure}[h]
    \centering
    \includegraphics[width=0.68\linewidth]{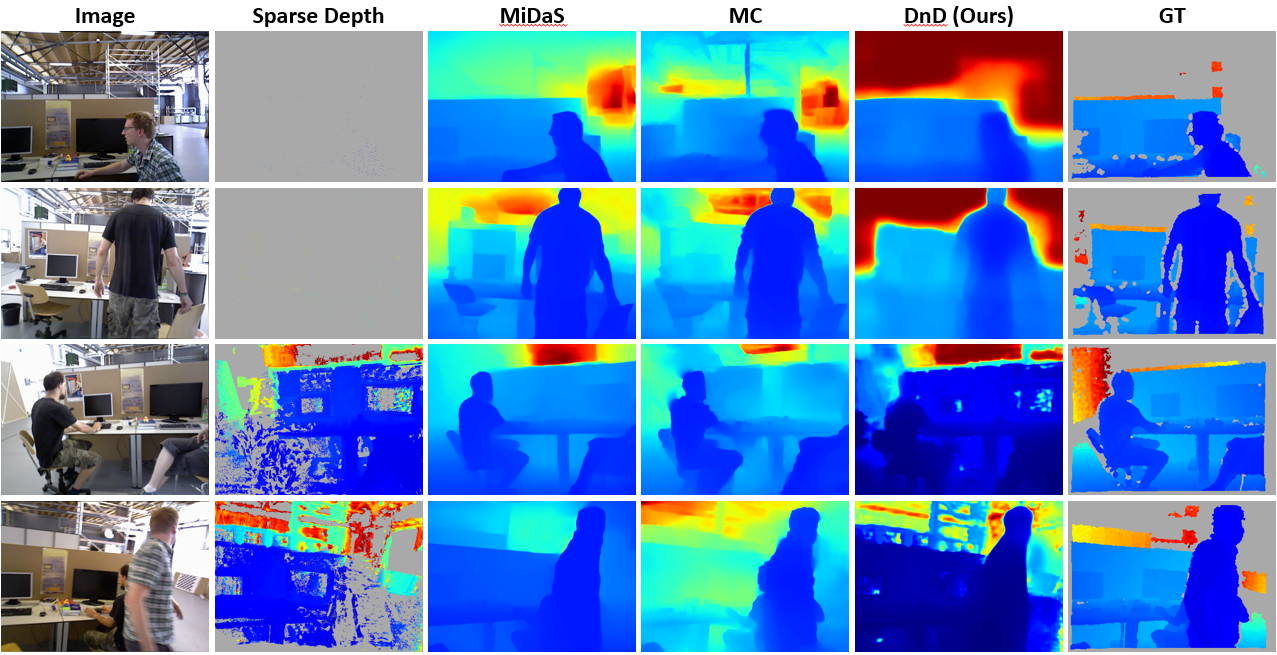}
    \vspace{-2mm}
    \caption{The qualitative results of the TUM RGB-D dataset. 
    The 1st and 2nd raw images result from the sparse depth maps, which are the sampled values of the Kinect with SIFT features. The 3rd and 4th raw images indicate the results for the projection of the scale ambiguous 3D model.
    Although there is no significant improvement in the visual quality compared to MiDaS \cite{ranftl2020towards} and Mannequin (MC) \cite{li2019learning}, DnD is able to estimate more accurate depth values in the numerical results since our method is based on the metric depth.
    }
    \label{fig:TUM RGB-D}
\end{figure}

Figure \ref{fig:nyu} and \ref{fig:TUM RGB-D} show the qualitative results of NYUv2 and  TUM RGB-D, respectively.
We acquire the sparse depth map by sparsifying the groundtruth depth map in order to obtain the metric scale.
Assuming the similar situation that the reprojected depth maps are derived from the 3D model, we only sample the depth values in the position of SIFT features for the corresponding images. In TUM, we used the ground truth camera pose provided by the dataset. In NYUv2, we used R2D2 \cite{r2d2} to obtain correspondences by using FLANN-based search algorithm and estimate the relative pose via the Perspective-n-Point (PnP \cite{lepetit2009epnp}).
\vspace{-2mm}

Since the dynamic objects category of the TUM RGB-D dataset is a small-scale dataset, this dataset is not suitable for self-supervised training method like DnD. As shown in Fig. \ref{fig:TUM RGB-D}, DnD does not show improved performance on qualitative results compared to MiDaS and Mannequin (MC). Both MiDaS and MC are based on supervised learning which require dense ground truth depth maps across different environments. These works trained their model on diverse and large-scale datasets. Specifically, MC used 136K image-depth pairs from Mannequin challenge videos, and MiDaS was trained on 10 different large-scale datasets \footnote{https://github.com/intel-isl/MiDaS}. However, DnD was trained on either MS or Dept datasets (25K or 60K images) without ground truth depth maps. Therefore, DnD does not show better generalization ability to produce sharp depth maps. Instead, since we focus on building a novel method to estimate metric depth maps, our method can show consistently better performance on quantitative results on TUM RGB-D dataset.        

\newpage
\section{Additional Qualitative Comparisons}
In this section, we provide more additional qualitative comparisons (See Figure 4. in the manuscript).
\begin{figure*}[h]
    \centering
    \includegraphics[width=0.86\linewidth]{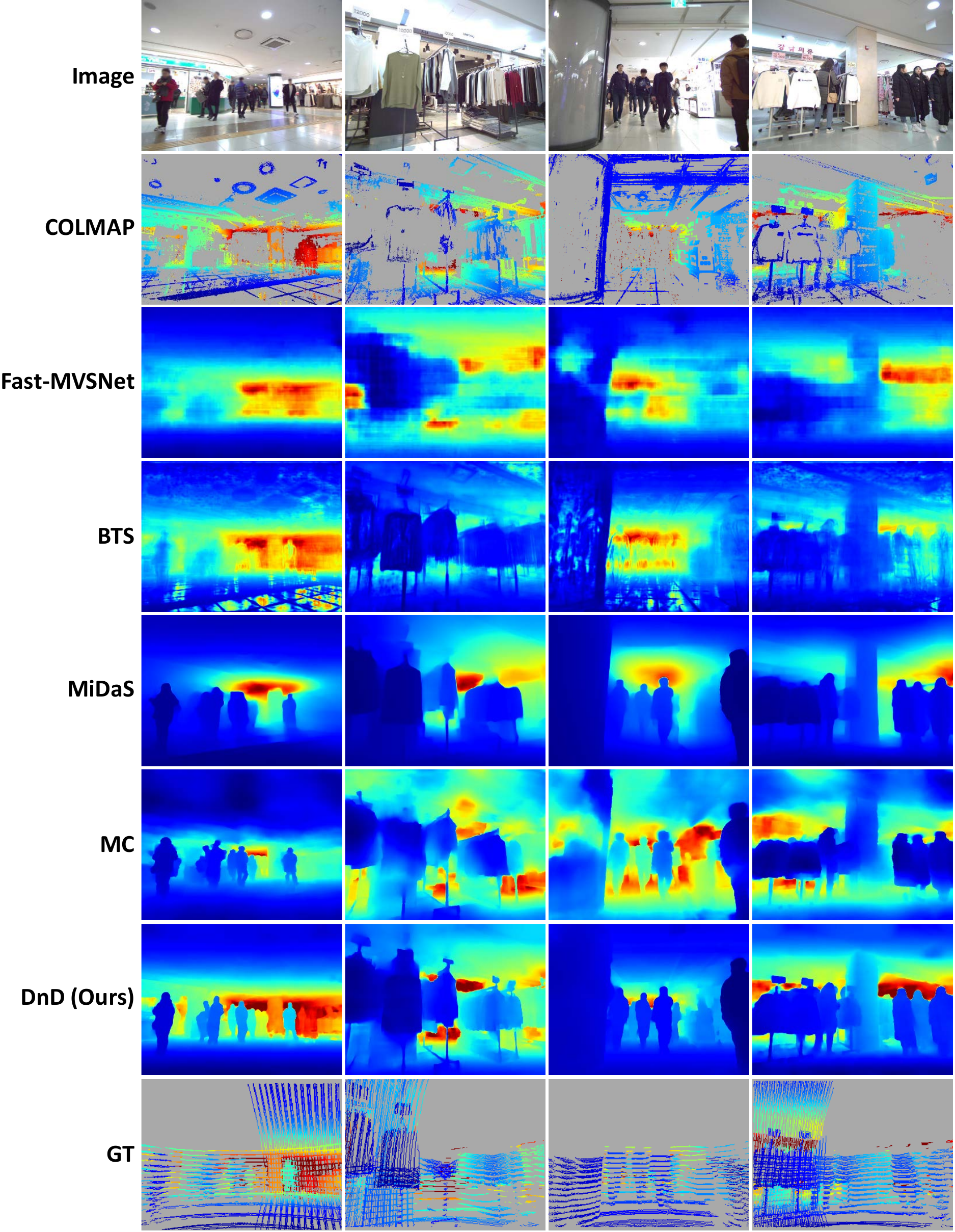}
    \caption{Qualitative comparisons with the state-of-the-art depth estimation algorithms in the MS dataset. From the third row to bottom, the results of Fast-MVSNet \cite{yu2020fast}, BTS \cite{lee2019big}, MiDaS \cite{ranftl2020towards}, Mannequin (MC) \cite{li2019learning}, and DnD are shown. All color maps use the jet color map (low \includegraphics[width=0.5cm,height=0.2cm]{figures/colorbar.jpeg} high; grey means empty depth values.)}
    \label{fig:qual_GN}
\end{figure*}
\begin{figure*}[h]
    \centering
    \includegraphics[width=0.86\linewidth]{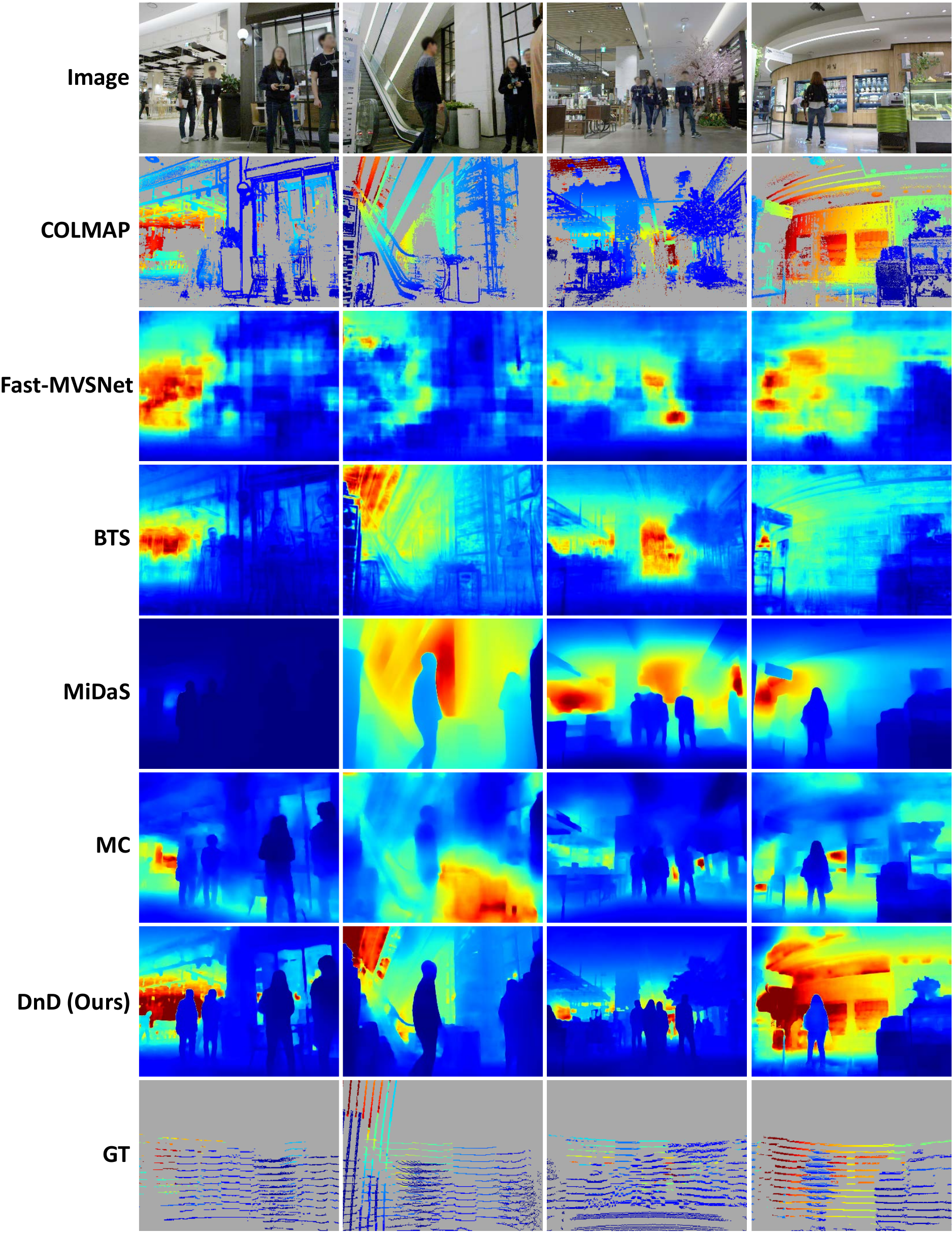}
    \caption{Qualitative comparisons with the state-of-the-art depth estimation algorithms in the Dept dataset. From the third row to bottom, the results of Fast-MVSNet \cite{yu2020fast}, BTS \cite{lee2019big}, MiDaS \cite{ranftl2020towards}, Mannequin (MC) \cite{li2019learning}, and DnD are shown. All color maps use the jet color map (low \includegraphics[width=0.5cm,height=0.2cm]{figures/colorbar.jpeg} high; grey means empty depth values.)}
    \label{fig:qual_HD}
\end{figure*}
\clearpage
\newpage

{\small
\bibliographystyle{ieee_fullname}
\bibliography{main_supp}
}